\documentclass[lettersize,journal]{IEEEtran}
\usepackage{amsmath,amsfonts}
\usepackage{algorithmic}
\usepackage{algorithm}
\usepackage{array}
\usepackage[caption=false,font=normalsize,labelfont=sf,textfont=sf]{subfig}
\usepackage{textcomp}
\usepackage{stfloats}
\usepackage{url}
\usepackage{verbatim}
\usepackage{graphicx}
\usepackage{cite}
\usepackage{dashbox} 
\usepackage{orcidlink}
\usepackage{cleveref}
\crefname{section}{§}{§§}
\Crefname{section}{§}{§§}
\crefname{table}{Table}{Tables}
\Crefname{table}{Table}{Tables}
\crefname{figure}{Figure}{Figures}
\Crefname{figure}{Figure}{Figures}
\crefname{equation}{Equation}{Equations}
\Crefname{equation}{Equation}{Equations}
\crefname{algorithm}{Algorithm}{Algorithms}
\Crefname{algorithm}{Algorithm}{Algorithms}

\usepackage{enumitem}
\usepackage{booktabs}
\usepackage{multirow}
\usepackage{threeparttable}
\usepackage{colortbl}
\usepackage{pifont}
\usepackage{array}
\usepackage{xcolor}
\usepackage{subcaption}

\usepackage{dashbox}

\newcommand{\name}{NetMamba+}

\definecolor{shapecolor}{rgb}{0.6,0.1,0.0}

\definecolor{commentcolor}{rgb}{0.5,0.5,0.5}

\begin{document}

\title{\name{}: A Framework of Pre-trained Models for Efficient and Accurate Network Traffic Classification}

\author{
Tongze Wang\orcidlink{0009-0007-0052-8402},
Xiaohui Xie\orcidlink{0000-0001-9413-4461},~\IEEEmembership{Member,~IEEE,}
Wenduo Wang\orcidlink{0009-0001-9929-6336},
Chuyi Wang\orcidlink{0009-0002-8296-3785},
Jinzhou Liu\orcidlink{0009-0006-4195-9623},
Boyan Huang\orcidlink{0009-0004-8035-4497},
Yannan Hu\orcidlink{0009-0008-3960-553X},
Youjian Zhao\orcidlink{0000-0001-9841-1796},
Yong Cui\orcidlink{0000-0002-5171-739X},~\IEEEmembership{Member,~IEEE}
\thanks{This work is supported by the NSFC Project under Grant 62132009, Grant 62221003 and Grant 62394322. An earlier version of this paper [arxiv link: https://arxiv.org/abs/2405.11449v3] has been accepted by the 32nd IEEE International Conference on Network Protocols (ICNP 2024). (Corresponding authors: Xiaohui Xie and Yong Cui.)}
\thanks{Tongze Wang is with the Institute for Network Sciences and Cyberspace, Tsinghua University, Beijing 100084, China.}
\thanks{Xiaohui Xie, Wenduo Wang, Chuyi Wang, Youjian Zhao, and Yong Cui are with the Department of Computer Science and Technology, Tsinghua University, Beijing 100084, China(email: xiexiaohui@mail.tsinghua.edu.cn; cuiyong@tsinghua.edu.cn).}
\thanks{Jinzhou Liu and Yannan Hu are with the Zhongguancun Laboratory, Beijing, China.}
\thanks{Boyan Huang is with the Central South University, Changsha, China.}
}

\markboth{Journal of \LaTeX\ Class Files,~Vol.~14, No.~8, August~2021}%
{Shell \MakeLowercase{\textit{et al.}}: A Sample Article Using IEEEtran.cls for IEEE Journals}


\maketitle

\begin{abstract}
With the rapid growth of encrypted network traffic, effective traffic classification has become essential for network security and quality of service management. Current machine learning and deep learning approaches for traffic classification face three critical challenges: computational inefficiency of Transformer architectures, inadequate traffic representations with loss of crucial byte-level features while retaining detrimental biases, and poor handling of long-tail distributions in real-world data. 
We propose \name{}, a framework that addresses these challenges through three key innovations: (1) an efficient architecture considering Mamba and Flash Attention mechanisms, (2) a multimodal traffic representation scheme that preserves essential traffic information while eliminating biases, and (3) a label distribution-aware fine-tuning strategy.
Evaluation experiments on massive datasets encompassing four main classification tasks showcase \name{}'s superior classification performance compared to state-of-the-art baselines, with improvements of up to 6.44\% in F1 score. 
Moreover, \name{} demonstrates excellent efficiency, achieving 1.7× higher inference throughput than the best baseline while maintaining comparably low memory usage. Furthermore, \name{} exhibits superior few-shot learning abilities, achieving better classification performance with fewer labeled data. 
Additionally, we implement an online traffic classification system that demonstrates robust real-world performance with a throughput of 261.87 Mb/s.
As the first framework to adapt Mamba architecture for network traffic classification, \name{} opens new possibilities for efficient and accurate traffic analysis in complex network environments.
\end{abstract}


\section{Introduction}
\IEEEPARstart{N}{etwork} traffic classification, which aims to identify potential threats within traffic or classify the category of traffic originating from different applications or services, has become an increasingly vital research area.
This is crucial for ensuring cybersecurity, improving service quality and user experience, and enabling efficient network management. 
However, the widespread adoption of encryption techniques (e.g., TLS) and anonymous network technologies (e.g., VPN, Tor) has made the accurate analysis of complex traffic more challenging.

Researchers have proposed numerous approaches to address this issue, showing promising results yet facing severe limitations.
Conventional machine learning methods~\cite{hayes2016k, van2020flowprint, taylor2017robust}, primarily relying on manually engineered features or statistical attributes, often fail to capture accurate traffic representations due to the absence of raw traffic data. 
In contrast, deep learning approaches~\cite{liu2019fs, lotfollahi2020deep, zhang2023tfe} automatically extract features from raw byte-level data, leading to enhanced traffic classification capabilities. 
Nonetheless, these deep learning methods necessitate extensive labeled datasets, rendering the models susceptible to biases and impeding their adaptability to novel data distributions.

Recently, pre-training has emerged as a prevalent model training paradigm in natural language processing~(NLP)~\cite{devlin2018bert} and computer vision~(CV)~\cite{he2022masked}. 
Motivated by this trend, several Transformer-based pre-trained traffic models~\cite{lin2022bert, zhao2023yet, wang2024lens} have been developed to learn generic traffic representations from extensive unlabeled data and then fine-tune for specific downstream tasks using limited labeled traffic data. However, these existing models face three significant challenges: 1) Limited Model Efficiency: state-of-the-art methods in traffic analysis primarily use Transformer architecture, which employs a quadratic self-attention mechanism to calculate correlations within a sequence. This leads to substantial computational and memory costs on long sequences\cite{zhu2024vision, qu2024trafficgpt}. Consequently, these models are unsuitable for real-time online traffic classification and cannot operate efficiently with the limited resources of typical network devices. 2) Inadequate Traffic Representation: current methods fail to capture traffic data accurately, often discarding essential byte- or transmission-level information while retaining unwanted biases. As a result, these unreliable schemes impair classification performance or even cause model failure in complex traffic scenarios. 3) Suboptimal Fine-tuning Strategy: existing approaches overlook the inherent class imbalance in traffic data, which commonly follow a long-tailed distribution. This oversight prevents models from achieving optimal performance in practical deployment.

To address these challenges, we propose \name{}, a framework for pre-trained models that incorporates an efficient underlying architecture, a comprehensive traffic representation scheme, and label distribution-aware fine-tuning. The goal is to accurately perform network traffic classification tasks with improved inference speed and reduced memory usage.

To improve model efficiency, we use the Mamba architecture or Flash Attention for the model backbone instead of the vanilla Transformer.
Mamba\cite{gu2023mamba}, a liner-time state space model for sequence modeling, has achieved notable success across various domains, including natural language processing~\cite{he2024densemamba}, computer vision~\cite{zhu2024vision} and graph understanding~\cite{wang2024graph}. 
This suggests promising potential for applying Mamba to the network domain. 
By carefully testing different variants of Mamba, we found that the unidirectional Mamba\cite{gu2023mamba} equipped with a residual connection, without omnidirectional scans or redundant blocks, is well-suited for efficiently learning latent patterns within sequential network traffic. 

Alternatively, Flash Attention\cite{dao2022flashattention} has been introduced to accelerate quadratic attention mechanisms through IO-aware techniques. To further enhance classification performance, we optimize the vanilla Transformer by integrating Flash Attention for improved efficiency, a pre-normalization architecture for greater stability, and a GeGLU-activated feedforward network~(FFN) to boost accuracy.

For traffic representation, we design a multimodal scheme that preserves valuable packet content from both headers and payloads, captures critical transmission patterns, and mitigates unwanted biases through techniques such as packet anonymization, byte allocation balancing, and stride-based data cutting, thereby enhancing traffic classification performance.

Finally, to better handle imbalanced datasets, we propose a novel label distribution-aware fine-tuning strategy. This approach reduces the negative effects of long-tailed distributions by assigning higher weights and enforcing larger margins for minority classes.

Specifically, \name{} initially extracts multimodal flow features from raw traffic and integrates cross-modal information through embedding. 
Subsequently, \name{} undergoes self-supervised pre-training on large unlabeled datasets, which is designed to learn generic representations of traffic data through reconstructing masked strides, zeroed packet sizes, and zeroed inter-arrival times. 
Finally, the decoder is replaced with a multi-layer perceptron head, and \name{} is fine-tuned on limited labeled data to refine traffic representations and adapt to downstream traffic classification tasks.

Extensive experiments conducted on publicly available datasets demonstrate the effectiveness and efficiency of \name{}. In all classification tasks, \name{} achieves best or near-best accuracy, with improvements of up to 6.44\% in F1 score. Compared to existing baselines, it attains 1.7× higher inference throughput while maintaining low GPU memory usage. Moreover, \name{} exhibits superior few-shot learning capabilities in comparison to other pre-training models, achieving better performance with fewer labeled data. In four out-of-distribution tasks, \name{} achieves AUROC scores exceeding 94\%. Additionally, we implement a \name{}-based online traffic classification system, which achieves an average throughput of 261.87 Mb/s, validating its practical effectiveness in real-world deployment.

In summary, our work makes the following contributions:
\begin{enumerate}[label=(\arabic*)]
    \item We propose \name{}, the first framework incorporating a state space model and Flash Attention-based modern Transformer specifically designed for network traffic classification(\Cref{sec:framework} and \Cref{sec:detailed-design}). Compared to existing Transformer-based methods, \name{} demonstrates superior performance and inference efficiency. Additionally, the code of \name{} is publicly available~\footnote{\url{https://github.com/wangtz19/NetMamba}}. 
    \item We develop a multimodal representation scheme for network traffic data that preserves valuable traffic characteristics while eliminating unwanted biases(\Cref{sec:traffic-representation}).
    \item We propose a novel label distribution-aware fine-tuning strategy to address data imbalance and improve model performance(\Cref{sec:flowmamba-finetuning}). 
    \item We implement an online system for \name{}, demonstrating its effectiveness in real-world deployment(\Cref{sec:online-sys}).
\end{enumerate}
\section{Related Work}
\subsection{Transformer-based Traffic Classification}
Due to its highly parallel architecture and robust sequence modeling abilities, Transformer has gained significant popularity and is extensively used for traffic understanding and generation tasks. For instance, MTT~\cite{zheng2022mtt} employs a multi‑task Transformer trained on truncated packet byte sequences to analyze traffic features in a supervised way. Recognizing the challenges associated with data annotation, MT-FlowFormer~\cite{zhao2022mt} introduces a Transformer-based semi-supervised framework for data augmentation and model improvement. 

To leverage unlabeled data effectively, several pre-trained models have been proposed. Inspired by BERT's pre-training methodology in natural language processing, PERT~\cite{he2020pert} and ET-BERT~\cite{lin2022bert} process raw traffic bytes using tokenization, apply masked language modeling to learn traffic representations, and fine-tune the models for downstream tasks. Similarly, YaTC~\cite{zhao2023yet} and FlowMAE~\cite{hang2023flow} adopt the widely-used MAE pre-training approach from computer vision, which involves patch splitting for byte matrices, capturing traffic correlations through masked patch reconstruction, and subsequent fine-tuning. 

Given the global interest in large language models, pre-trained traffic foundation models such as NetGPT~\cite{meng2023netgpt} and Lens~\cite{wang2024lens} have been developed to address traffic analysis and generation simultaneously.
However, Transformer-based models face computational and memory inefficiencies because of the quadratic complexity of their core self-attention mechanism. This necessitates a more efficient and effective solution for online traffic classification.
We advance existing work by optimizing the vallina Transformer, by incorporating Flash Attention to enhance efficiency, a pre-normalization architecture to improve stability, and a GeGLU-activated feedforward network to further boost accuracy.

\subsection{Mamba-based Representation Learning}
Representation learning is a branch of machine learning concerned with automatically learning and extracting meaningful representations or features from raw data. Since the advent of Mamba, an efficient and effective sequence model, numerous Mamba variants have emerged to enhance representation learning across diverse domain-specific data formats. For instance, in the realm of vision tasks requiring spatial awareness, custom-designed scan architectures like Vim~\cite{zhu2024vision} and VMamba~\cite{liu2024vmamba} have been developed. In the domain of language modeling, DenseMamba~\cite{he2024densemamba} improves upon the original SSM by incorporating dense internal connections to boost performance. Handling graph data necessitates specialized solutions such as Graph-Mamba~\cite{wang2024graph} and STG-Mamba~\cite{li2024stg}, each employing tailored graph-specific selection mechanisms. Furthermore, various Mamba variants have proven effective in domains like signal processing~\cite{li2024spmamba}, point cloud analysis~\cite{liang2024pointmamba}, and multi-modal learning~\cite{qiao2024vl}.

However, to date, there are no reports of Mamba's successful application in network traffic classification, highlighting the need for our research in this area.
We push the boundaries of current approaches by optimizing unidirectional Mamba for network traffic classification, providing both an efficient and accurate solution.

\subsection{Traffic Representation Schemes}

In real-world scenarios, massive raw network traffic encompasses a wide range of data categories that vary in upper applications, carried protocols, or transmission purposes. Therefore, a robust representation scheme with appropriate granularity is crucial for accurate traffic understanding.

Traditional machine learning methods~\cite{hayes2016k, van2020flowprint, taylor2017robust, barradas2021flowlens, zhou2023efficient}, constrained by limited model parameters and fitting capabilities, commonly resort to utilizing compressed statistical features at the packet or flow level, such as distributions of packet sizes or inter-arrival times. However, these features often suffer from excessive compression, resulting in the loss of vital information inherent in raw datagrams. 

Recent advancements in deep learning have endeavored to utilize raw traffic bytes. However, as shown in~\cref{tb:rep-comp}, these methods face limitations. They often neglect crucial information in packet headers or transmission patterns and introduce unwanted biases by ignoring byte balance or using improper data-splitting techniques.

To address these issues, we propose a novel network traffic representation scheme. 
Our approach remedies the aforementioned shortcomings, preserving hierarchical traffic information while effectively eliminating biases.  

\begin{table}[ht]
    \footnotesize
    \centering
    \caption{Comparison of Existing Representation Schemes}
    \begin{threeparttable}
    \begin{tabular}{cccccc}
        \toprule
         Method & Header & Payload & BB\tnote{\P} & MM\tnote{*} &  Splitting \\
        \midrule
         PERT\cite{he2020pert} & \ding{55} & \ding{51} & \ding{55} & \ding{55} & token \\
         ET-BERT\cite{lin2022bert} & \ding{55} & \ding{51} & \ding{55} & \ding{55} & token \\
         YaTC\cite{zhao2023yet} & \ding{51} & \ding{51} & \ding{51} & \ding{55} & patch \\
         FlowMAE\cite{hang2023flow} & \ding{51} & \ding{51} & \ding{55} & \ding{55} & patch \\
         NetGPT\cite{meng2023netgpt} & \ding{51} & \ding{51} & \ding{55} & \ding{55} & token \\
         Lens\cite{wang2024lens} & \ding{51} & \ding{51} & \ding{55} & \ding{55} & token \\
         \midrule
         \textbf{NetMamba} & \ding{51} & \ding{51} & \ding{51} & \ding{55} & stride \\
         \textbf{\name{}} & \ding{51} & \ding{51} & \ding{51} & \ding{51} & stride \\
        \bottomrule
    \end{tabular}
    \begin{tablenotes}
        \item $^{\P}$ BB: Byte Balance sets fixed sizes for headers and payloads.
        \item $*$ MM: Multimodal input feautres.
    \end{tablenotes}
    \end{threeparttable}
    \label{tb:rep-comp}
\end{table}

\subsection{Deep Long-Tailed Learning}
Deep long-tailed learning addresses the challenge of training a deep neural network on datasets characterized by a highly imbalanced class distribution, where a small proportion of classes contain the majority of samples, while the remaining classes have only a few samples.

Existing methods for addressing long-tailed class imbalance can be broadly categorized into three main approaches: class re-balancing, information augmentation and module improvement \cite{zhang2023deep}. Among them, class re-balancing is a dominant approach that seeks to mitigate the negative impact of class imbalance during training. For example, re-sampling methods adjust the number of training samples per class based on the sample size \cite{kang2019decoupling} or class-wise training accuracy\cite{feng2021exploring}. To avoid the additional computational costs of oversampling, class-sensitive learning modifies the training loss for different classes by adapting loss weights \cite{cui2019class} or prediction probabilities\cite{cao2019learning} according to label frequencies. As a post-training solution, logit adjustment aims to calibrate the biased model predictions during the inference stage\cite{menon2020long} \cite{tian2020posterior}.

Building upon prior work, we introduce a label distribution-aware loss that assigns higher weights and enforces larger margins for minority classes. This approach mitigates the negative effects of long-tailed distribution without explicitly increasing training costs, and has been shown effective for traffic classification.
\section{Preliminaries}
This section elaborates on basic definitions, terminologies, and components underlying the Mamba 
block which serves as the foundation of the proposed \name{}. 
\subsubsection{State Space Models}
As the key components of Mamba, State Space Models~(SSMs) represent a contemporary category of sequence models within deep learning that share broad connections with Recurrent Neural Networks~(RNNs) and Convolutional Neural Networks~(CNNs). Drawing inspiration from continuous systems, SSMs are commonly structured as linear Ordinary Differential Equations~(ODEs) which establish a mapping from an input sequence $x(t) \in \mathbb{R}^{N}$ to an output sequence $y(t) \in \mathbb{R}^{N}$ via an intermediate latent state $h(t) \in \mathbb{R}^{N}$:

\begin{equation}
    \label{eq:ssm-continuous}
    \begin{aligned}
        h^{'}(t) & =\mathbf{A}h(t) + \mathbf{B}x(t) \\
        y(t) & = \mathbf{C}h(t)
    \end{aligned}
\end{equation}
where $\mathbf{A} \in \mathbb{R}^{N\times N}$ represents the evolution parameter, while $\mathbf{B} \in \mathbb{R}^{N\times 1}$ and $\mathbf{C} \in \mathbb{R}^{1\times N}$ are the projection parameters.

\subsubsection{Discretization}
Integrating raw SSMs with deep learning presents a significant challenge due to the discrete nature of typical real-world data, contrasting with the continuous-time characteristic of SSMs. To overcome this challenge, the zero-order hold (ZOH) technique is utilized for discretization, leading to the discrete version formulated as follows:

\begin{equation}
\label{eq:ssm-recurrent}
\begin{aligned}
h_{t} & = \mathbf{\overline{A}}h_{t-1} + \mathbf{\overline{B}}x_{t} \\
y_{t} & = \mathbf{C}h_{t}
\end{aligned}
\end{equation}
where $\mathbf{\overline{A}} = \exp(\Delta \mathbf{A})$ and 
$\mathbf{\overline{B}} \approx \Delta \mathbf{B}$ 
represent the discretized parameters, with $\Delta$ denoting the discretization step size. 
This recurrent formulation, known for its linear time complexity, is suitable for model inference but lacks parallelizability during training.


\subsubsection{Selective Scan}
 While designed for sequence modeling, SSMs exhibit subpar performance when content-aware reasoning is required, primarily due to their time-invariant nature. Specifically, the parameters $\mathbf{\overline{A}}$, $\mathbf{\overline{B}}$, and $\mathbf{C}$ remain constant across all input tokens within a sequence. To address this issue, Mamba~\cite{gu2023mamba} introduces the selection mechanism, enabling the model to select pertinent information from the context dynamically. This adaptation involves transforming the SSM parameters $\mathbf{\overline{B}}$, $\mathbf{C}$, and $\Delta$ into functions of the input $x$. 
 Moreover, to avoid the sequential recurrent computation outlined in \Cref{eq:ssm-recurrent}, Mamba adopts a work-efficient parallel scan algorithm~\cite{blelloch1990prefix, smith2022simplified}.
 A GPU-friendly implementation is also developed to efficiently compute the selection mechanism, leading to a notable reduction in memory I/O operations and eliminating the need to store intermediate states.

\section{\name{} Framework\label{sec:framework}}

\begin{figure*}[tp]
\centering
\includegraphics[scale=0.6]{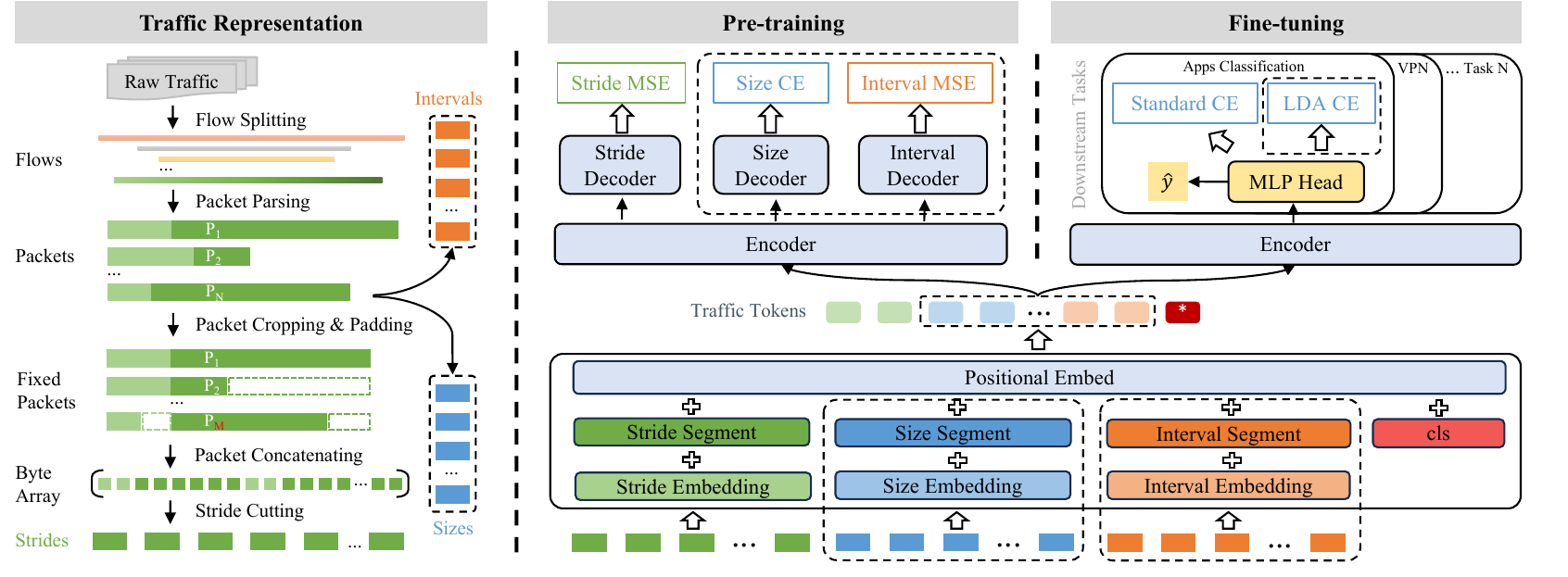}
\scriptsize
\textit{Pluggable components are enclosed within \dbox{\textbf{dashed boxes}}}
\caption{Overview of \name{} Framework.}
\label{fig:system-design}
\end{figure*}

This section overviews the framework of \name{}~(see \cref{fig:system-design}), providing a comprehensive blueprint for the detailed design presented in \cref{sec:traffic-representation} and \cref{sec:detailed-design}. Initially, \name{} extracts hierarchical information from raw binary traffic and converts it into multimodal representation. Inspired by the Masked AutoEncoders~(MAE) pre-training model in computer vision, \name{} employs a dual-stage training approach. Specifically, self-supervised pre-training is utilized to acquire traffic representation, while supervised fine-tuning is employed to tailor the model for downstream traffic understanding tasks.

\subsubsection{Traffic Representation Phase}
To enhance domain knowledge within networks, \name{} adopts a multimodal methodology to represent key content within network traffic. 
Initially, traffic data is segmented into distinct flows, categorized by their 5-tuple attributes: Source IP, Destination IP, Source Port, Destination Port, and Protocol. 
Fixed-sized segments of header and payload bytes are then extracted for each packet within a flow. 
To collect more comprehensive traffic information without compromising model efficiency due to excessively long packet sequences, we follow approaches outlined in prior studies\cite{lin2022bert, zhao2023yet}, which involve selectively utilizing specific packets within a flow. 
Specifically, bytes from the initial packets of each flow are aggregated into a unified byte array, integrating information across byte, packet, and flow levels for a comprehensive view of traffic characteristics.

This byte array forms the foundation for segmenting non-overlapping flow strides. It preserves semantic relationships between adjacent bytes, effectively mitigating biases introduced by conventional patch-splitting methods, as well as addressing out-of-vocabulary issues commonly associated with tokenization processes. 

To incorporate transmission patterns not captured by raw bytes, we introduce an additional modality by extracting sequences of packet sizes and inter-arrival times, which encode critical spatial and temporal characteristics of each connection.
More design intricacies regarding traffic representation are elucidated in \cref{sec:traffic-representation}.

\subsubsection{Pre-training Phase}
To acquire generic encodings of network domain knowledge based on flow stride representations, \name{} undergoes pre-training using extensive unlabeled network traffic data.
Specifically, \name{} utilizes a masked autoencoder~(MAE) architecture, incorporating multiple unidirectional Mamba blocks in both its encoder and decoder, as detailed in \cref{sec:mamba-block}. 

During pre-training, traffic strides undergo several sequential steps: concatenation with a trailing class token, mapping into stride embeddings, addition of positional embeddings, and random masking. 
The encoder focuses solely on visible strides, grasping inherent relationships and generating an output traffic representation. 
The decoder then reconstructs the masked strides using the encoder's output and dummy tokens. 

For packet sizes and inter-arrival times, tokens are randomly zeroed before entering the encoder and then reconstructed by the decoder. Unlike masked strides, both zeroed and non-zeroed tokens are visible to the encoder.
Pre-training is optimized by minimizing the reconstruction loss for the masked strides, ensuring the model learns robust traffic patterns.
Detailed insights into the pre-training strategy are provided in \cref{sec:flowmamba-pretraining}.

\subsubsection{Fine-tuning Phase}
For accurately capturing traffic patterns and understanding downstream task requirements, \name{} undergoes fine-tuning using labeled traffic data.
During this phase, the decoder of \name{} is replaced by a multi-layer perceptron~(MLP) head to facilitate classification tasks. 
With the removal of the reconstruction task, all embedded flow strides become visible to the encoder. 
Subsequently, \name{} forwards only the trailing class token to the MLP-based classifier.
Specifically, we propose a pluggable label distribution-aware fine-tuning strategy to adapt \name{} for long-tailed datasets.

Post pre-training, \name{}'s encoder exhibits significant adaptability when fine-tuned with limited labeled data, enabling efficient transition to various downstream tasks such as application classification and attack detection. Additional details on the label distribution-aware fine-tuning process are provided in \cref{sec:flowmamba-finetuning}.
\section{Traffic Representation \label{sec:traffic-representation}}
This section provides detailed information about the traffic representation scheme used by \name{}. 

\subsubsection{Flow Splitting}
Formally, given network traffic comprising multiple packets, we segment it into various flows, with each flow consisting of packets that belong to a specific protocol and are transmitted between two ports on two hosts. Packets within the same flow encapsulate significant interaction information between the two hosts. This information includes the establishment of a TCP connection, data exchanged during communication, and the overall transmission status. These flow-level features are pivotal in characterizing application behaviors and enhancing the efficiency of traffic classification processes.

\subsubsection{Packet Parsing}
For each flow, all packets are processed through several sequential operations to preserve valuable information and eliminate unnecessary interference. 
When narrowing down the scope for analyzing traffic data related to specific applications or services, we exclude all packets carried by non-IP protocols, such as Address Resolution Protocol~(ARP) and Dynamic Host Configuration Protocol~(DHCP). 
Considering the critical information contained within both the packet (e.g., the total length field) and the payload (text content for upper-level protocols), we choose to retain these elements.
Furthermore, to mitigate biases introduced by identifiable information, all packets are anonymized through the removal of 
Ethernet headers 
and the masking of IP addresses and ports.

\subsubsection{Packet Cropping \& Padding, and Concatenating}
Given the variability in packet size within the same flow and the fluctuation in both header length (including the IP header and any 
upper-layer headers) and payload length within individual packets, problematic scenarios often arise. For instance, the first long packet can occupy the entire limited model input array, or excessively long payloads can dominate the byte information within shorter headers. Therefore, it is essential to standardize packet sizes by assigning uniform sizes to all packets and fixed lengths to both packet headers and payloads.
Specifically, we select the first $M_b$ packets from a single flow, setting the header length to $N_h$ bytes and the payload length to $N_p$ bytes. 
Any packet exceeding this length will be cropped, while shorter packets will be padded to meet these specifications. 

Eventually, all bytes of initial $M_b$ packets are concatenated into a unified array $[b_1, b_2, \dots, b_{L_b}]$ where $L_b = M_b \times (N_h + N_p)$ represents the array length and $b_i$ denotes the $i$-th byte. 

\subsubsection{Stride Cutting}
Given the significant computational and memory demands posed by a byte array with $L_b$ (typically greater than 1000) elements, it becomes imperative to explore further compression techniques to enhance the efficiency of model training and inference. Traditional methods often involve reshaping the byte array into a square matrix and employing two-dimensional patch splitting, a practice borrowed from computer vision. However, this technique unintentionally introduces biases by grouping vertically adjacent bytes that are semantically unrelated, as they are not naturally contiguous in the sequential traffic data.

Inspired by patching methods used in time-series forecasting, we adopt a 1-dimensional stride cutting approach on the original array, aligning with the sequential nature of network traffic and preserving inter-byte correlations. 
Specifically, we divide the byte array into non-overlapping strides of size $1 \times L_s$, resulting in a total number of strides $N_{\text{stride}} = L_b / L_s$. Each stride $\mathbf{s}_i \in \mathbb{R}^{1 \times L_s}$ is defined as $[b_{L_s \times i}, b_{L_s \times i + 1}, \dots, b_{L_s \times (i+1) - 1}]$ for $0 \le i < N_{\text{stride}}$. 
This strategy aims to mitigate biases while retaining essential sequential information in the data.

\subsubsection{Pluggable Sequence Extraction} Although raw bytes inherently encode rich traffic characteristics, input length constraints limit the model to the first $M_b$ packets, inevitably leading to the loss of global transmission patterns. To compensate for this limitation, we introduce an additional feature modality—sequential features (i.e., sequences of packet sizes and inter-arrival times)—into \name{}. Specifically, we extract sizes and intervals of the first $M_{\text{seq}}$ packets ($M_{\text{seq}} > M_b$).

To mitigate numerical instability arising from the wide value range of sequential features, we apply tailored normalization strategies. For the packet size sequence, we perform clamping to ensure that each value does not exceed the maximum transmission unit (MTU), formulated as:
$$
x_{\text{size}} = \min\{x_{\text{size}}, \mathtt{MTU}\}.
$$
For the packet interval sequence, we first reduce distributional skewness through logarithmic scaling, followed by a sigmoid transformation to map the values into the range $[0,1]$. The normalization is defined as:
$$
x_{\text{int}} = \text{sigmoid}(\log(x_{\text{int}})) = \frac{1}{1 + 1/(1+x_{\text{int}})}.
$$

\textbf{Takeaway}. \emph{Our traffic representation scheme effectively retains crucial information from both packet headers and payloads, while eliminating unwanted biases through techniques such as IP and port removal, byte balancing, and stride cutting. In addition, we incorporate packet size and interval sequences to capture essential transmission patterns. For a detailed evaluation, please refer to \cref{sec:ab-study} and \cref{sec:ab-mmf}.}
\section{Model Details \label{sec:detailed-design}}
This section details the \name{} model architecture, along with the pre-training and fine-tuning strategies.

\subsection{Embedding Layer}
\subsubsection{Stride Embedding}
Given the stride array, we initially perform a linear projection on  each stride $\mathbf{s}_i$ to a vector with size $\mathtt{D}_{\text{enc}}$ and incorporate position embeddings $\mathbf{PE}_{\text{enc}}$ as shown below:
\begin{equation}
\begin{aligned}
    \mathbf{X}_{\text{stride}} &=  \left[ \mathbf{s}_1\mathbf{W}; \mathbf{s}_2\mathbf{W}; \cdots; \mathbf{s}_{N_{\text{stride}}}\mathbf{W}; \mathbf{x}_{\text{cls}}\right] \\
    \mathbf{X}_0 &= \mathbf{X}_{\text{stride}} + \mathbf{PE}_{\text{enc}}
\end{aligned}
\end{equation}
where $\mathbf{W} \in \mathbb{R}^{L_s \times \mathtt{D}_{\text{enc}}}$ represents the learnable projection matrix. Inspired by ViT~\cite{dosovitskiy2020image} and BERT~\cite{devlin2018bert}, we introduce a class token to represent the entire stride sequence, denoted as $\mathbf{x}_{\text{cls}}$. Since the unidirectional Mamba processes sequence information from front to back, we opt to append the class token to the end of the sequence for enhanced information aggregation.

\subsubsection{Pluggable Multimodal Embedding}
To ensure that our model fully captures cross-modal correlations, we adopt an early-fusion mechanism that relies on multimodal embedding.
The embedding of traffic strides is implemented via a linear transformation. While this approach is applicable to size and interval sequences, it fails to capture relative variations and temporal trends in the latent embedding space. Inspired by the sinusoidal positional encoding in Transformers, we apply fixed sinusoidal encoding to both size and interval sequences in order to preserve their sequential dependencies in the embedding space. Formally, for a given size or interval sequence $\mathbf{x}$ of length $M_{\text{seq}}$, the encoding is defined as:
$$
\mathbf{SE}(\mathbf{x}) = \left[ \mathbf{SE}(x_1), \mathbf{SE}(x_2), \cdots, \mathbf{SE}(x_{M_{\text{seq}}}) \right]
$$
where each individual element is mapped as:
$$
\begin{cases}
\mathbf{SE}_{(x_i, 2j)} &=  \sin(x_i / 10000^{2j/\mathtt{D_{enc}}}) \\
\mathbf{SE}_{(x_i, 2j+1)} &=  \cos(x_i / 10000^{2j/\mathtt{D_{enc}}})
\end{cases}
$$

To integrate stride embeddings $\mathbf{X}_{\text{stride}} \in \mathbb{R}^{N_{\text{stride}} \times \mathtt{D_{enc}}}$, size embeddings $\mathbf{X}_{\text{size}} \in \mathbb{R}^{M_{\text{seq}} \times \mathtt{D_{enc}}}$, and interval embeddings $\mathbf{X}_{\text{int}} \in \mathbb{R}^{M_{\text{seq}} \times \mathtt{D_{enc}}}$, we first add modality-specific segment indicators to distinguish feature sources. The embeddings are then concatenated together with an additional class token for global information aggregation. Subsequently, learnable positional embeddings are added to preserve positional information, and the resulting representations are transformed into traffic tokens. The fusion process can be formulated as:
$$
\mathbf{X}_0 = \left[\mathbf{X}_{\text{stride}} + \mathbf{I}_{\text{stride}};\, \mathbf{X}_{\text{size}} + \mathbf{I}_{\text{size}};\, \mathbf{X}_{\text{int}} + \mathbf{I}_{\text{int}};\, \mathbf{x}_{\text{cls}}\right] + \mathbf{PE}_{\text{enc}}
$$

\subsection{Encoder \& Decoder\label{sec:mamba-block}}

Both the \name{} encoder and decoder consist of multiple sequence-modeling blocks, the detailed processes of each block type are outlined below.

\subsubsection{NetMamba Block}
Recently, several variants of Mamba have been proposed to accommodate domain-specific data formats and task requirements. For instance, Vim~\cite{zhu2024vision} incorporates bidirectional Mamba blocks for spatial-aware understanding of vision tasks, Graph-Mamba~\cite{wang2024graph} introduces a graph-dependent selection mechanism for graph learning, while MiM-ISTD~\cite{chen2024mim} customizes a cascading Mamba structure for extracting hierarchical visual information. We argue that the original unidirectional Mamba \cite{gu2023mamba}, enhanced with an outer residual connection, is well-suited for representation learning in sequential network traffic. This configuration offers increased efficiency through the elimination of omnidirectional scans and redundant blocks. We carefully test different Mamba variants, demonstrating that the selected unidirectional Mamba is more suitable for processing network traffic. Please refer to the ablation studies for more details.

Hence, we implement the Mamba-based encoder and decoder using NetMamba blocks.
The operational process of the NetMamba block forward pass is outlined in \cref{alg:block}. For a given input token sequence $\mathbf{X}_{t-1}$ with a batch size $\mathtt{B}$ and sequence length $\mathtt{L}$ from the $(t-1)$-th NetMamba block, we begin by normalizing it and then projecting it linearly into $\mathbf{x}$ and $\mathbf{z}$, both with dimension size of $\mathtt{E}$. We subsequently apply causal 1-D convolution to $\mathbf{x}$, resulting in $\mathbf{x}'$. Based on $\mathbf{x}'$, we compute the input-dependent step size $\mathbf{\Delta}$, as well as the projection parameters $\mathbf{B}$ and $\mathbf{C}$ having a dimension size of $\mathtt{N}$. We then discretize $\overline{\mathbf{A}}$ and $\overline{\mathbf{B}}$ using $\mathbf{\Delta}$. Following this, we calculate $\mathbf{y}$ employing a hardware-aware SSM. Finally, $\mathbf{y}$ is gated by $\mathbf{z}$ and added residually to $\mathbf{X}_{t-1}$, resulting in the output token sequence $\mathbf{X}_{t}$ for the $t$-th NetMamba block. 

\begin{algorithm}[ht]
\caption{NetMamba Block Forward Pass}
\small
\begin{algorithmic}[1]
\REQUIRE{$\mathbf{X}_{t-1}$ : \textcolor{shapecolor}{$(\mathtt{B}, \mathtt{L}, \mathtt{D})$}}
\ENSURE{$\mathbf{X}_{t}$ : \textcolor{shapecolor}{$(\mathtt{B}, \mathtt{L}, \mathtt{D})$}}
\STATE $\mathbf{X}_{t-1}'$ : \textcolor{shapecolor}{$(\mathtt{B}, \mathtt{L}, \mathtt{D})$} $\leftarrow$ $\mathbf{Norm}(\mathbf{X}_{t-1})$ \COMMENT{normalize input sequence}
\STATE $\mathbf{x}$ : \textcolor{shapecolor}{$(\mathtt{B}, \mathtt{L}, \mathtt{E})$} $\leftarrow$ $\mathbf{Linear}^\mathbf{x}(\mathbf{X}_{t-1}')$ 
\STATE $\mathbf{z}$ : \textcolor{shapecolor}{$(\mathtt{B}, \mathtt{L}, \mathtt{E})$} $\leftarrow$ $\mathbf{Linear}^\mathbf{z}(\mathbf{X}_{t-1}')$
\STATE $\mathbf{x}'$ : \textcolor{shapecolor}{$(\mathtt{B}, \mathtt{L}, \mathtt{E})$} $\leftarrow$ $\mathbf{SiLU}(\mathbf{Conv1d}(\mathbf{x}))$
\STATE $\mathbf{B}$ : \textcolor{shapecolor}{$(\mathtt{B}, \mathtt{L}, \mathtt{N})$} $\leftarrow$ $\mathbf{Linear}^{\mathbf{B}}(\mathbf{x}')$ \COMMENT{input-dependent}
\STATE $\mathbf{C}$ : \textcolor{shapecolor}{$(\mathtt{B}, \mathtt{L}, \mathtt{N})$} $\leftarrow$ $\mathbf{Linear}^{\mathbf{C}}(\mathbf{x}')$ \COMMENT{input-dependent}
\STATE $\mathbf{\Delta}$ : \textcolor{shapecolor}{$(\mathtt{B}, \mathtt{L}, \mathtt{E})$} $\leftarrow$ $\log(1 + \exp(\mathbf{Linear}^{\mathbf{\Delta}}(\mathbf{x}') + \mathbf{Parameter}^{\mathbf{\Delta}}))$ \COMMENT{softplus ensures positive step size, input-dependent}
\STATE $\overline{\mathbf{A}}$ : \textcolor{shapecolor}{$(\mathtt{B}, \mathtt{L}, \mathtt{E}, \mathtt{N})$} $\leftarrow$ $\mathbf{\Delta} \bigotimes \mathbf{Parameter}^{\mathbf{A}}$ \COMMENT{discritize}
\STATE $\overline{\mathbf{B}}$ : \textcolor{shapecolor}{$(\mathtt{B}, \mathtt{L}, \mathtt{E}, \mathtt{N})$} $\leftarrow$ $\mathbf{\Delta} \bigotimes \mathbf{B}$ \COMMENT{discritize}
\STATE $\mathbf{y}$ : \textcolor{shapecolor}{$(\mathtt{B}, \mathtt{L}, \mathtt{E})$} $\leftarrow$ $\mathbf{SSM}(\overline{\mathbf{A}}, \overline{\mathbf{B}}, \mathbf{C})(\mathbf{x}')$ \COMMENT{hardware-aware scan}
\STATE $\mathbf{y}'$ : \textcolor{shapecolor}{$(\mathtt{B}, \mathtt{L}, \mathtt{E})$} $\leftarrow$ $\mathbf{y} \bigodot \mathbf{SiLU}(\mathbf{z}) $ \COMMENT{self-gating}
\STATE $\mathbf{X}_{t}$ : \textcolor{shapecolor}{$(\mathtt{B}, \mathtt{L}, \mathtt{D})$} $\leftarrow$ $\mathbf{Linear}^\mathbf{X}(\mathbf{y}') + \mathbf{X}_{t-1}$ \COMMENT{residual connection}
\STATE Return: $\mathbf{X}_{t}$ \COMMENT{output sequence}
\end{algorithmic}
\label{alg:block}
\end{algorithm}

\subsubsection{NetTrans Block} 
To overcome the bottleneck of vanilla Transformers in scaling to long sequences, numerous efficient Transformer variants have been proposed, including linear Transformers\cite{katharopoulos2020transformers, choromanski2020rethinking, wang2020linformer} with $O(\mathtt{L})$ complexity and sparse Transformers\cite{kitaev2020reformer, beltagy2020longformer, zaheer2020big} with $O(\mathtt{L}\log\mathtt{L})$ complexity. 
These methods achieve speed-up by approximating attention, reducing the original $O(\mathtt{L}^2)$ complexity of vanilla attention. However, this comes at the cost of reduced  accuracy due to over-compression. A significant advancement in the area, Flash Attention, accelerates the quadratic attention process through reducing memory accesses and decreases memory consumption by discarding intermediate values. Thanks to its high efficiency, Flash Attention has become a core component of modern Transformer-based models\cite{warner2024smarter, ahdritz2024openfold}.

In light of this, we implement a NetTrans block featuring Flash Attention 2 for improved efficiency, a pre-normalized architecture for more stable training\cite{xiong2020layer}, and a GeGLU-activated feedfarward network~(FFN) for higher accuracy\cite{shazeer2020glu}. 
For a token sequence $\mathbf{X}_{t-1}$, the following steps are performed:
\begin{enumerate}[label=\arabic*.]
\item The sequence pass through the attention module to capture inter-token dependencies:
\begin{equation}
\begin{aligned}
    \mathbf{X}_{t-1}^{1} &= \textbf{LayerNorm}(\mathbf{X}_{t-1}) \\
    \mathbf{X}_{t-1}^{2} &= \textbf{FlashAttention}(\mathbf{X}_{t-1}^{1}) \\
    \mathbf{X}_{t-1}^{3} &= \mathbf{X}_{t-1}^{1} + \mathbf{X}_{t-1}^{2}
\end{aligned}
\end{equation}
\item Next, inter-dimension correlations are computed within the FFN module, yielding the NetTrans block's output:
\begin{equation}
\begin{aligned}
    \mathbf{X}_{t-1}^{4} &= \textbf{LayerNorm}(\mathbf{X}_{t-1}^{3}) \\
    \mathbf{X}_{t-1}^{5} &= \textbf{FFN}_{\textbf{GeGLU}}(\mathbf{X}_{t-1}^{4}) \\
    \mathbf{X}_{t} &= \mathbf{X}_{t-1}^{4} + \mathbf{X}_{t-1}^{5}
\end{aligned}
\end{equation}
\end{enumerate}

\subsection{\name{} Pre-training \label{sec:flowmamba-pretraining}}
\subsubsection{Stride Pre-training}
Given the embedded stride tokens $\mathbf{X}_0 \in \mathbb{R}^{\mathtt{L}\times \mathtt{D}_{\text{enc}}}$, a portion of strides is randomly sampled while the remaining ones are removed.
For a predefined masking ratio, let the number of visible tokens be denoted as $\mathtt{L}_{\text{vis}}$. The visible tokens are obtained as follows:
\begin{equation}
    \mathbf{X}_0^{\text{vis}} = \mathbf{Shuffle}(\mathbf{X}_0)[1:\mathtt{L}_{\text{vis}}, \ :\ ]
    \in \mathbb{R}^{\mathtt{L}_{\text{vis}}\times \mathtt{D}_{\text{enc}}}
\end{equation}
where the $\mathbf{Shuffle}$ operation permutes the token sequence randomly. 
Notably, we ensure that the trailing class token remains unmasked throughout this process since its role in aggregating overall sequence information necessitates its preservation at all times.

The primary objective behind random masking is the elimination of redundancy. This approach creates a challenging task that resists straightforward solutions through extrapolation from neighboring strides alone. 
Additionally, the reduction in input length diminishes computational and memory costs, offering an opportunity for more efficient model training. 

The encoder is tasked with capturing latent inter-stride relationships using the visible tokens, whereas the stride decoder's objective is to reconstruct masked strides utilizing both the encoder output tokens and mask tokens. Each mask token represents a shared, trainable vector indicating the presence of a missing stride. Additionally, new positional embeddings are added to provide location information to the mask tokens.

The formal forward process of \name{} pre-training can be outlined as follows:
\begin{equation}
\begin{aligned}
    \mathbf{X}_{\text{enc}}^{\text{out}} & = \mathbf{MLP}(\mathbf{Encoder}(\mathbf{X}_0^{\text{vis}})) \in \mathbb{R}^{\mathtt{L}_{\text{vis}}\times \mathtt{D}_{\text{dec}}}\\
    \mathbf{X}_{\text{dec}}^{\text{in}} & = \mathbf{Unshuffle}(\mathbf{Concat}(\mathbf{X}_{\text{enc}}^{\text{out}}, \mathbf{X}_{\text{mask}})) + \mathbf{PE}_{\text{dec}} \\
    \mathbf{X}_{\text{dec}}^{\text{out}} & = \mathbf{Decoder}(\mathbf{X}_{\text{dec}}^{\text{in}})
\end{aligned}
\end{equation}
where the $\mathbf{Unshuffle}$ operation restores the original sequence order, and $\mathbf{PE}_{\text{dec}} \in \mathbb{R}^{\mathtt{L}\times \mathtt{D}_{\text{dec}}}$ represents decoder-specific positional embeddings. Subsequently, the mean square error~(MSE) loss for self-supervised reconstruction is calculated as shown below:
\begin{equation}
\begin{aligned}
    \mathbf{x}_{\text{stride-masked}} & = \mathbf{Shuffle}(\mathbf{X}_0)[\mathtt{L}_{\text{vis}}+1:\mathtt{L}, \ :\ ] \\
    \hat{\mathbf{x}}_{\text{stride-masked}} & = \mathbf{Shuffle}(\mathbf{X}_{\text{dec}}^{\text{out}})[\mathtt{L}_{\text{vis}}+1:\mathtt{L}, \ :\ ] \\
    \mathcal{L}_{\text{stride-rec}} & = \mathbf{MSE}(\mathbf{x}_{\text{stride-masked}}, \hat{\mathbf{x}}_{\text{stride-masked}})
\end{aligned}
\end{equation}
where $\mathbf{x}_{\text{stride-masked}}$ represents the ground-truth mask tokens, and $\hat{\mathbf{x}}_{\text{stride-masked}}$ signifies the predicted ones.

\subsubsection{Pluggable Multimodal Pre-training}
For pre-training in the multimodal setting, we retain the MAE-based pre-training scheme for traffic strides, and introduce additional strategies for size and interval sequences. Specifically, we randomly mask a subset of sequence tokens by setting them to zero, feed the corrupted sequences into the encoder, and reconstruct the masked values using modality-specific decoders. 

Since packet sizes take discrete values, their reconstruction is formulated as a classification problem, with the objective defined by the cross-entropy loss:
$$
\mathcal{L}_{\text{size-rec}} = \mathbf{CE}(\mathbf{x}_{\text{size-zeroed}}, \hat{\mathbf{x}}_{\text{size-zeroed}}).
$$
In contrast, packet intervals lie in a continuous and unbounded range, so their reconstruction is treated as a regression problem, optimized via the mean squared error (MSE) loss:
$$
\mathcal{L}_{\text{int-rec}} = \mathbf{MSE}(\mathbf{x}_{\text{int-zeroed}}, \hat{\mathbf{x}}_{\text{int-zeroed}}).
$$
The overall multimodal pre-training objective is thus given by:
$$
\mathcal{L}_{\text{rec}} = \mathcal{L}_{\text{stride-rec}} + \mathcal{L}_{\text{size-rec}} + \mathcal{L}_{\text{int-rec}}.
$$

\subsection{\name{} Fine-tuning \label{sec:flowmamba-finetuning}}
\subsubsection{Normal Fine-tuning}
For downstream tasks, all encoder parameters, including embedding modules and Mamba blocks, are loaded from pre-training. To conduct classification on labeled traffic data, the decoder is replaced with an MLP head. Given that all tokens are visible, fine-tuning of \name{} is performed in a supervised manner as detailed below:
\begin{equation}
\begin{aligned}
    \mathbf{X} & = \mathbf{Encoder}(\mathbf{X}_0) \in \mathbb{R}^{\mathtt{L}\times \mathtt{D}_{\text{enc}}} \\
    \mathbf{z} & = \mathbf{MLP(\mathbf{Norm}(\mathbf{X}[\mathtt{L}, \ :]))} \in \mathbb{R}^{\mathtt{C}}
\end{aligned}
\end{equation}
Here, $\mathbf{z}$ represents the prediction logits, where $\mathtt{C}$ is the number of traffic categories. For simplicity, we denote the $j$-th prediction logit as $z_j$ for $j \in \{1, \dots, \mathtt{C} \}$. Given the ground-truth one-hot label $\mathbf{y} \in \mathbb{R}^{\mathtt{C}}$, which satisfies $y_y = 1$ and $y_j = 0$ for $j \neq y$, i.e., the ground-truth label is $y$, the standard cross-entropy loss is computed as follows:
\begin{equation}
\mathcal{L}_{\text{CE}} = -\sum_{i=1}^{\mathtt{C}}y_i\log\frac{e^{z_i}}{\sum_{j=1}^{\mathtt{C}}e^{z_j}} = -\log\frac{e^{z_y}}{\sum_{j=1}^{\mathtt{C}}e^{z_j}}
\end{equation}

\subsubsection{Pluggable Label Distribution-Aware Fine-tuning}
To adapt the model training for imbalanced datasets, the Class-Balanced~(CB) loss \cite{cui2019class} re-weights the loss values inversely proportional to the effective number of samples, thereby giving more importance to minority classes. Let the training sample size of class $j$ be denoted as $n_j$ and $\beta \in [0,1)$ be a hyper-parameter, the class-balanced cross-entropy loss is given by:
\begin{equation}
\mathcal{L}_{\text{CB}} = -\frac{1-\beta}{1-\beta^{n_y}} \log \left( \frac{e^{z_y}}{\sum_{j=1}^{\mathtt{C}}e^{z_j}} \right)
\end{equation}
Here, $\beta=0$ corresponds to no re-weighting, and as $\beta \rightarrow 1$, the re-weighting is based on the inverse class frequency.

To further enhance generalization for minority classes, the Label Distribution-Aware Margin~(LDAM)\cite{cao2019learning} enforces larger margins for these classes. Given a hyper-parameter $C$, the LDAM-based cross-entropy loss is written as:
\begin{equation}
\mathcal{L}_{\text{LDAM}} = -\log \left( \frac{e^{z_y - \Delta_y}}{e^{z_y - \Delta_y} + \sum_{j\neq y}e^{z_j}} \right)
\end{equation}
where $\Delta_j = C/n_j^{1/4}$ for $j \in \{1, \dots, \mathtt{C} \}$.

Combining both the re-weighting and re-margining strategies, we propose a novel label distribution-aware~(LDA) cross-entropy loss that allocates more weight and enforces larger margins for minority classes:
\begin{equation}
\mathcal{L}_{\text{LDA}} = -\frac{1-\beta}{1-\beta^{n_y}} \log \left( \frac{e^{z_y - \Delta_y}}{e^{z_y - \Delta_y} + \sum_{j\neq y}e^{z_j}} \right)
\end{equation}

\textbf{Takeaway}. \emph{To integrate multimodal features, we employ early fusion via multimodal embedding. To improve model efficiency, we implement the NetMamba block based on unidirectional Mamba and the NetTrans block using Flash Attention. To acquire generic network domain knowledge, \name{} is pre-trained by reconstructing masked strides, zeroed packet sizes and zeroed inter-arrival times. For downstream adaptation, \name{} is fine-tuned with either the standard cross-entropy loss or a new label distribution-aware loss for handling long-tailed datasets.}

\section{Online System Implementation\label{sec:online-sys}}
\begin{figure}[tp]
\centering
\includegraphics[scale=0.3]{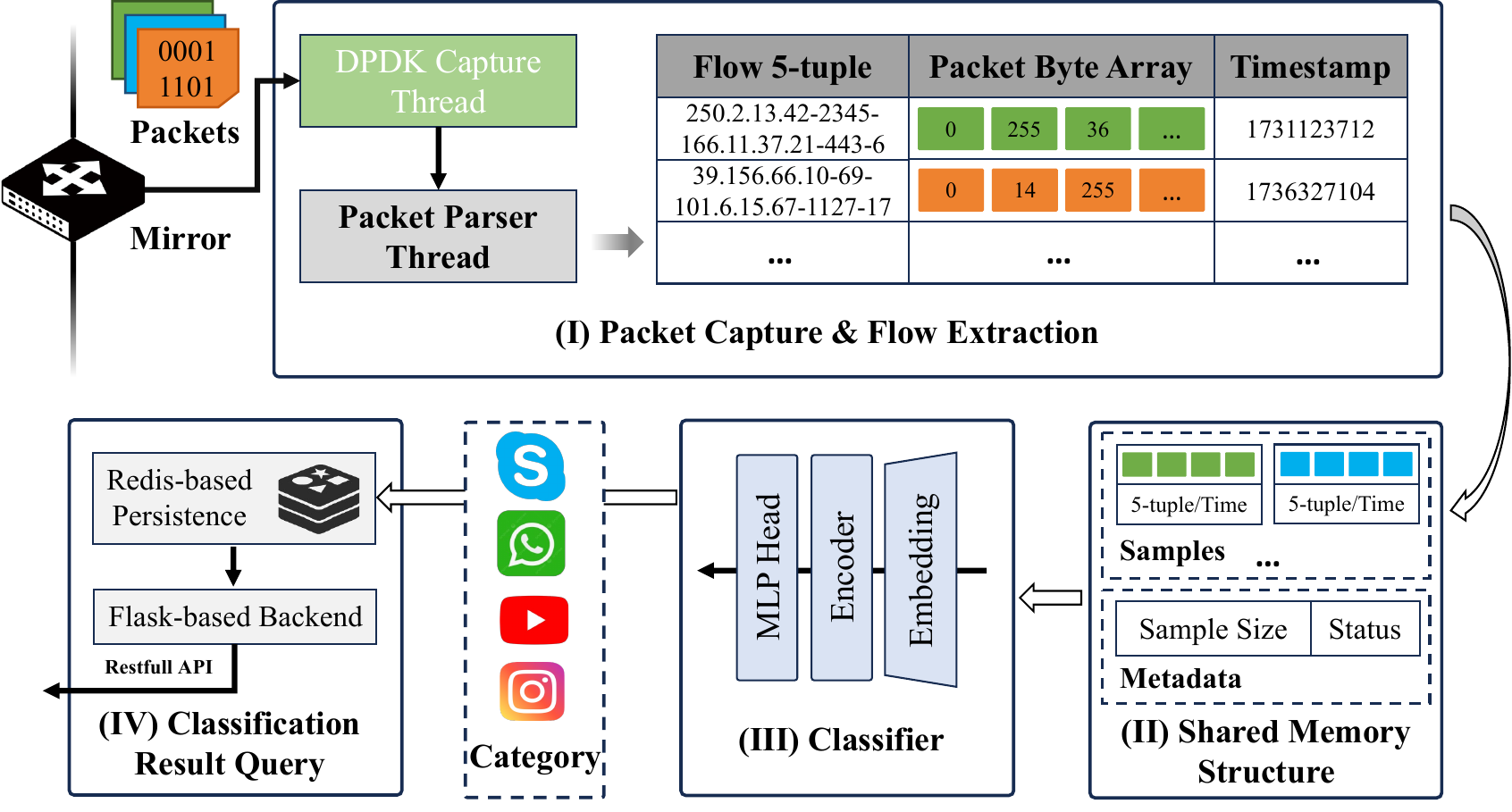}
\caption{The Prototype of NetMamba-based Online Traffic Classification System.}
\label{fig:proto-sys}
\end{figure}

To enable NetMamba for online inference in real-world environments, we implement a prototype system as shown in \Cref{fig:proto-sys}. This system captures network packets using the Data Plane Development Kit~(DPDK)\cite{dpdk} within a CPU process\footnote{DPDK is a set of libraries and drivers designed to accelerate packet processing in user space, bypassing the kernel networking stack for higher performance.}, stores the processed flow samples in shared memory, runs the trained NetMamba classifier in a GPU process, and persists the classification results into Redis for subsequent querying.

\subsection{Packet Capture \& Flow Extraction}
This module captures high-speed network traffic and extracts specific features for each flow. 
After being mirrored into the system, packets are first captured by DPDK in one thread, parsed in another thread, and finally buffered in a flow table.
The flow table contains three fields: the \textit{flow 5-tuple}, which serves as an identifier, the \textit{packet byte array}, which buffers incoming packet bytes, and the \textit{timestamp}, which is used for periodic memory management. 

Initially, the flow table is empty.
Upon receiving a packet, its first $N_h$ header bytes and $N_p$ payload bytes are extracted and stored in the corresponding packet byte array according to its 5-tuple.
If a flow 5-tuple is not found in the flow table, a new entry is created, and its timestamp is set to the time of the first packet arrival. 

Every $W_g$ seconds, the entire flow table is traversed,  and any entry that contains at least $M_b$ packets is transferred to the shared memory. To prevent short flows from overwhelming the memory, each flow entry is removed $W_r$ seconds after its creation, regardless of the number of packets it contains.

\subsection{Shared Memory, Classifier \& Result Query}
Shared memory is used to store incoming flow samples for classification. It holds samples consisting of flow bytes and the 5-tuple, as well as metadata including \textit{sample size}, which indicates the total number samples, and \textit{status}, which indicates whether data has been consumed by the classifier.

When the extraction process writes new data into shared memory, it updates \textit{sample size} and sets \textit{status} to 1. The classifier process continuously monitors shared memory. Upon detecting new data (i.e. when \textit{status} equals 1), it fetches all samples based on \textit{size}, resets \textit{status} to 0, and then performs classification.

The newly generated flow categories are persisted in a Redis database for future result queries. Additionally, we provide RESTful APIs by running a Flask backend to facilitate query operation.

\section{Evaluation}
\subsection{Experimental Setup}
\subsubsection{Datasets}
\begin{table*}[ht]
    \footnotesize
    \centering
    \caption{Statistics of Used Datasets}
    \begin{threeparttable}
    \begin{tabular}{ccccccccc}
        \toprule
         Stage & Dataset & Downstream Task & Protocols$^{\P}$ & EFR$^{*}$ & EPR$^{\dagger}$ & MFC$^{\ddagger}$ & Flows & Categories \\ 
        \midrule
        Pre-training & Browser~\cite{van2020flowprint} & - & TLS1.3, TLS1.2, GQUIC... & 0.5162 & 0.3038 & $\infty$ & 149528 & - \\
        Pre-training & Kitsune~\cite{mirsky2018kitsune} & - & TLS1.1, SSDP... & 0.0025 & 0.1351 & $\infty$ & 167831 & - \\
        \midrule
         Fine-tuning & CipherSpectrum~\cite{wickramasinghe2025sok} & Application Classification & TLS1.3, TLS1.1 & 1.0000 & 0.3643 & 2000 & 82000 & 41 \\
         Fine-tuning & CSTNET-TLS1.3~\cite{lin2022bert} & Application Classification & TLS1.3, SSLv2, TLS1.2...  & 1.0000 & 0.3447 & 2000 & 92705 & 119 \\
         Fine-tuning & CrossNet2021A~\cite{li2022prograph} & Application Classification & TLS1.2, TLS1.3, HTTP...  & 0.6037 & 0.2360 & 2000 & 8843 & 19 \\
         Fine-tuning & CP-Android~\cite{ren2019international} & Application Classification & TLS1.2, HTTP, TLS1.1...  & 0.2043 & 0.2223 & 2000 & 17219 & 179 \\
         Fine-tuning & CP-iOS~\cite{ren2019international} & Application Classification & TLS1.2, TLS1.1, HTTP...  & 0.4107 & 0.3553 & 2000 & 9049 & 121 \\
         Fine-tuning & CICIoT2022~\cite{dadkhah2022towards} & Attack Classification & RTP, HTTP, TLS1.2... & 0.0022 & 0.0024 & 2000 & 10404 & 6 \\
         Fine-tuning & USTC-TFC2016~\cite{wang2017malware} & Malware Classification & NBSS, HTTP, FTP...  & 0.0532 & 0.0135 & 2000 & 4000 & 20 \\
         Fine-tuning & ISCXVPN2016~\cite{gil2016characterization} & VPN Classification & TLS1.2, SSHv2, STUN... & 0.0670 & 0.0524 & 2000 & 10135 & 7 \\
         Fine-tuning & DataCon2021-p1~\cite{datacon2021} & VPN Classification & TLS1.3, TLS1.1, WireGuard...  & 0.2852 & 0.1868 & 2000 & 3823 & 11 \\
         Fine-tuning & Huawei-VPN & VPN Classification & TLS1.2, QUIC, TLS1.3 & 0.5778 & 0.2333 & $\infty$ & 38704 & 12 \\
        \bottomrule
    \end{tabular}
    \begin{tablenotes}
        \item $^{\P}$ We report only special protocols that account for a substantial proportion of the traffic, while excluding common protocols such as IP, TCP, UDP, ICMP, ARP, and DNS. The reported protocols are ranked according to their relative proportions.
        \item $^{*}$EFR: Encrypted Flow Ratio; $^{\dagger}$ EPR: Encrypted Packet Ratio; $^{\ddagger}$MFC: Maximal Flows per Category.
    \end{tablenotes}
    \end{threeparttable}
    \label{tb:datasets}
\end{table*}

As shown in \cref{tb:datasets}, we use two distinct datasets for pre-training: Browser~\cite{van2020flowprint} and Kitsune~\cite{mirsky2018kitsune}. The Browser dataset contains encrypted application traffic collected on a Samsung smartphone while accessing the top 1,000 Alexa-ranked websites using different browsers. The Kitsune dataset consists of network traffic from various attacks (e.g., Denial of Service, Botnet, Man-in-the-Middle, Reconnaissance), captured either from a commercial surveillance system or from a network full of IoT devices. 

For fine-tuning, we utilize both public datasets and enterprise proprietary traffic data, covering four primary classification tasks.
\begin{enumerate}[label=\arabic*.]
    \item \textbf{Encrypted Application Classification}: This task focuses on identifying application traffic transmitted under different encryption protocols. 
    We evaluate on the CipherSpectrum~\cite{wickramasinghe2025sok}, CSTNET-TLS1.3~\cite{lin2022bert}, CrossNet2021A~\cite{li2022prograph}, CP-Android~\cite{ren2019international} and CP-iOS~\cite{ren2019international} datasets. Notably, most flows of CipherSpectrum and CSTNET-TLS1.3are encrypted with modern protocols such as TLS 1.3 and strong cipher suites (e.g.,  CHACHA20\_POLY1305).
    
    \item \textbf{Attack Traffic Classification}: This task aims to detect malicious traffic, including Denial-of-Service~(DoS) and brute-force attacks. We construct 6 data categories using CICIoT2022\cite{dadkhah2022towards}.
    
    \item \textbf{Malware Traffic Classification}: This task seeks to distinguish between traffic generated by malware and benign applications. We adopt all 20 traffic categories from the USTC-TFC2016 dataset \cite{wang2017malware}.
    
    \item \textbf{Encrypted VPN Classification}: This task involves identifying VPN protocols used by various applications. We use VPN traffic from 7 service categories in ISCXVPN2016~\cite{gil2016characterization}, together with proxy application traffic in DataCon2021-p1~\cite{datacon2021} and real-world traffic collected from the Huawei production campus.
    For data collection, we isolated the target VPN application on a mobile phone by closing all other applications and captured the corresponding traffic at an intercity router by storing packets from the device's IP address.
\end{enumerate}
To alleviate the issue of data imbalance, we impose an upper limit of flows per category for most fine-tuning dataset, as detailed in \cref{tb:datasets}. After filtering, each dataset is partitioned into training, validation, and test sets following an 8:1:1 ratio. 


\subsubsection{Comparison Methods}
To comprehensively evaluate \name{}, we conducted comparisons with various open-source baselines and state-of-the-art~(SOTA) techniques, as outlined below:
\begin{enumerate}[label=\arabic*.]
    \item Classical machine learning methods such as \textbf{AppScanner} \cite{taylor2017robust} and \textbf{FlowPrint} \cite{van2020flowprint} that rely on statistical features for traffic classification.
    \item Deep learning approaches like \textbf{Seq2Img}~\cite{chen2017seq2img}, \textbf{FS-Net}~\cite{liu2019fs}, \textbf{FlowPic}~\cite{shapira2019flowpic}, \textbf{mini-FlowPic}~\cite{horowicz2022few}
    and \textbf{TFE-GNN} \cite{zhang2023tfe} that utilize packet lengths or raw bytes to perform traffic analysis in a supervised manner.
    \item Transformer-based models such as \textbf{ET-BERT}~\cite{lin2022bert}, \textbf{YaTC}~\cite{zhao2023yet} and \textbf{TrafficFormer}~\cite{zhou2025trafficformer} that capture traffic representations during pre-training and subsequently fine-tune for specific tasks with limited labeled data. 
    \item Transformer variants within the \name{} backbone, including \textbf{NetTransV} and \textbf{NetTransL}. The former replaces all \name{} blocks with vanilla Transformer blocks \cite{vaswani2017attention} featuring quadratic complexity, while the latter adopts Linear Transformer blocks \cite{katharopoulos2020transformers} with linear complexity.
    \item Mamba variants within the \name{} backbone, including \textbf{NetMambaB} and \textbf{NetMambaC}. The former replaces all \name{} blocks with bi-directional Mamba blocks \cite{zhu2024vision}, while the latter uses cascading Mamba blocks \cite{chen2024mim}.
\end{enumerate} 

\subsubsection{Implementation Details}
We implement three variants of our models: \textbf{NetTrans}, \textbf{NetMamba} and \textbf{NetMamba+}. The former consists entirely of NetTrans blocks, while the latter two are composed solely of NetMamba blocks. Notably, both NetTrans and NetMamba process raw bytes only, while NetMamba+ includes multimodal inputs. The \name{} architecture includes an encoder composed of 4 blocks and a decoder with 2 blocks. More hyper-parameter details can be found in \cref{tb:hyper-param}. Unless otherwise specified, the LDA fine-tuning is disabled.

At the pre-training stage, we set the batch size to $\mathtt{B} = 128$ and train models for 150,000 steps. The initial learning rate is set to $1.0 \times 10^{-3}$ with the AdamW optimizer, alongside a linear learning rate scaling policy. Additionally, we apply a masking ratio of 0.9 to randomly mask strides and 0.15 to packet sizes and inter-arrival times.

For fine-tuning, we adjust the batch size to $\mathtt{B} = 64$ and set the learning rate to $2.0 \times 10^{-3}$. 
All models are trained for 120 epochs on the training data, with checkpoints saving the best accuracy on the validation set, subsequently evaluated on the test set.

The proposed model is implemented using PyTorch 2.1.1, with all offline experiments conducted on a Ubuntu 22.04 server equipped with CPU of Intel(R) Xeon(R) Gold 6240C CPU @ 2.60GHz, GPU of NVIDIA A100 (40GB $\times$ 4).

\begingroup
\renewcommand{\arraystretch}{1.3}
\begin{table}[ht]
    \footnotesize
    \centering
    \caption{Hyper-Parameter details of \name{}}
    \begin{tabular}{c|c||c|c||c|c}
        \toprule
        \hline
         Variable & Value & Variable & Value & Variable & Value \\ \hline
         $M_b$ & 5 & $\mathtt{D}_{\text{enc}}$ & 256 & $L_s$ & 4 \\ \hline
         $N_h$ & 80 & $\mathtt{D}_{\text{dec}}$ & 128 & $\mathtt{N}$ & 16 \\ \hline
         $N_p$ & 240 & $\mathtt{E}_{\text{enc}}$ & 512 & $\mathtt{L}$ & 401 \\ \hline
         $M_{\text{seq}}$ & 20 & $\mathtt{E}_{\text{dec}}$ & 256  & $\mathtt{L}_{\text{vis}}$ & 41 \\ \hline
        \bottomrule
    \end{tabular}
    \label{tb:hyper-param}
\end{table}
\endgroup

\subsubsection{Evaluation Metrics}
For in-distribution evaluation, we assess the performance of \name{} using four typical metrics: Accuracy(AC), Precision(PR), Recall(RC), and weighted F1 Score(F1). For out-of-distribution evaluation, we adopt the Area Under the Receiver Operating Characteristic curve (AUROC) and the False Positive Rate at 95\% True Positive Rate (FPR95). 

\subsection{Overall Evaluation}
\subsubsection{Comparison with SOTA Methods on Public Datasets}

\begin{table*}[htpb]
    \centering
    \caption{Comparison against State-of-the-art Methods}
    \begin{tabular}{c|cc|cc|cc|cc|cc|cc}
    \toprule
    \multirow{2}{*}{Method} & \multicolumn{2}{c|}{Params(M)} & \multicolumn{2}{c|}{CipherSpectrum} & \multicolumn{2}{c|}{CSTNET-TLS1.3} & \multicolumn{2}{c|}{CICIoT2022} & \multicolumn{2}{c|}{ISCXVPN2016} & \multicolumn{2}{c}{USTC-TFC2016}\\
    \cmidrule(lr){2-13} \noalign{\vskip 0.5mm}
    & PT & FT & AC & F1 & AC & F1 & AC & F1 & AC & F1 & AC & F1 \\
    \midrule
    AppScanner\cite{taylor2017robust} & - & - &
    0.4851 & 0.6357 & 0.1987 & 0.2980 & 0.5315 & 0.6884 & 0.4142 & 0.4863 & 0.4193 & 0.5310 \\
    FlowPrint\cite{van2020flowprint} & - & - &
    0.2488 & 0.2362 & 0.1264 & 0.1100 & 0.6901 & 0.6202 & 0.9251 & 0.9266 & 0.5614 & 0.5134 \\
    \midrule
    Seq2Img\cite{chen2017seq2img} & - & 16.9 &
    0.8513 & 0.8515 & 0.7383 & 0.7354 & 0.9447 & 0.9443 & 0.7757 & 0.7702 & 0.8241 & 0.8023 \\
    FS-Net\cite{liu2019fs} & - & 5.3 &
    \underline{0.9000} & \underline{0.9008} & \underline{0.7870} & \underline{0.7845} & 0.8252 & 0.8252 & 0.7731 & 0.7692 & 0.7450 & 0.7485 \\
    FlowPic\cite{shapira2019flowpic} & - & 0.06 &
    0.4421 & 0.4249 & 0.2732 & 0.2195 & 0.4678 & 0.4714 & 0.3281 & 0.3299 & 0.3185 & 0.3109 \\
    mini-FlowPic\cite{horowicz2022few} & 0.04 & 0.05 &
    0.4054 & 0.3981 & 0.0570 & 0.0307 & 0.3487 & 0.2705 & 0.2868 & 0.2573 & 0.2740 & 0.2591 \\
    TFE-GNN\cite{zhang2023tfe} & - & 44.3 &
    0.7573 & 0.7525 & 0.3386 & 0.3100 & \cellcolor{lightgray}0.9975 & \cellcolor{lightgray}0.9975 & 0.8342 & 0.8243 & 0.9685 & 0.9654 \\
    \midrule
    ET-BERT\cite{lin2022bert} & 187.4 & 136.4 &
    0.7026 & 0.7046 & 0.5002 & 0.4935 & 0.9481 & 0.9480 & 0.8900 & 0.8893 & 0.9735 & 0.9730 \\
    YaTC \cite{zhao2023yet} & 2.3 & 2.1 &
    0.8589 & 0.8577 & 0.7830 & 0.7793 & 0.9712 & 0.9712 & \underline{0.9411} & \underline{0.9414} & \underline{0.9795} & \underline{0.9793} \\
    TrafficFormer\cite{zhou2025trafficformer} & 187.4 & 136.4 &
    0.6071 & 0.6106 & 0.6674 & 0.6630 & 0.8893 & 0.8921 & 0.6907 & 0.6863 & 0.9593 & 0.9585 \\
    \midrule
    \textbf{NetTrans} & 3.1 & 2.4 &
    0.8543 & 0.8548 & 0.7654 & 0.7619 & 0.9769 & 0.9769 & \underline{0.9411} & 0.9408 & \cellcolor{lightgray}0.9823 & \cellcolor{lightgray}0.9822 \\
    \textbf{NetMamba} & 2.2 & 1.9 &
    0.8779 & 0.8783 & 0.7755 & 0.7728 & \underline{0.9779} & \underline{0.9779} & 0.9401 & 0.9401 & 0.9743 & 0.9740 \\
    \textbf{NetMamba+} & 2.6 & 1.9 &
    \cellcolor{lightgray}0.9652 & \cellcolor{lightgray}0.9652 & \cellcolor{lightgray}0.8498 & \cellcolor{lightgray}0.8489 & 0.9750 & 0.9750 & \cellcolor{lightgray}0.9460 & \cellcolor{lightgray}0.9460 & 0.9765 & 0.9765 \\
    \bottomrule
    \end{tabular}
    \label{tb:acc-sota}
\end{table*}

We evaluated the performance of our models in traffic classification using five publicly available datasets. As shown in \cref{tb:acc-sota}, NetMamba+, the only model with multimodal features, achieve the best performance among all baselines, while both NetMamba and NetTrans achieve comparable performance against SOTA methods.
On average, NetMamba+ achieves accuracies ranging from 0.8498 to 0.9765. Notably, our three models maintain a considerably low parameter count. This highlights the efficiency of our models in learning effective traffic representations.

For datasets composed entirely of encrypted flows, such as CipherSpectrum and CSTNET-TLS1.3, features extracted from raw bytes contain relative limited semantics since encrypted payloads contain little information. In such a scenario, FS-Net, which leverages packet size sequence features, achieves better classification performance than most deep or even pre-trained models. This underscores the importance of transmission patterns in encrypted traffic classification. Thanks to adopting multimodal input features, NetMamba+ achieves best performance by capturing byte-level and sequence-level correlations.

For datasets containing partially encrypted flows, such as CICIoT2022, ISCXVPN2016 and USTC-TFC2016, methods based on raw bytes (e.g., ET-BERT, YaTC, TrafficFormer, NetMamba, NetTrans) generally outperform those based on sequence representations (e.g., Seq2Img, FS-Net, FlowPic, mini-FlowPic). Specifically, our three models fall behind TFE-GNN on CICIoT2022, as TFE-GNN excludes all flows without payloads, while CICIoT2022 includes a large proportion of DoS traffic that contains only headers. Additionally, NetMamba+ ranks first on ISCXVPN2016 and  third on USTC-TFC2016, demonstrating its excellent performance in traffic classification.

\subsubsection{Comparison with Pre-trained Models on Real-world Dataset}

\begin{figure}[htpb]
\centering
\includegraphics[scale=0.4]{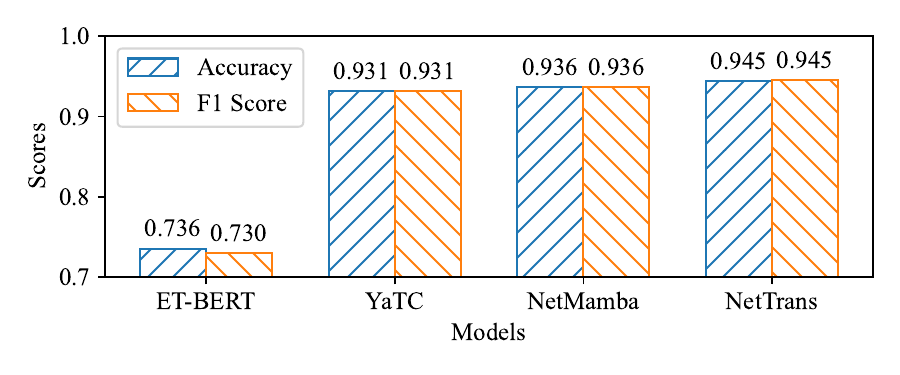}
\caption{Comparison of Pre-trained Models on Huawei-VPN Dataset}
\label{fig:acc-huawei}
\end{figure}
As observed from \cref{tb:acc-sota}, pre-trained models generally outperform the non-pre-trained models. For simplicity, we therefore focus the comparison on our models and pre-trained baselines using Huawei-VPN dataset. This dataset contains application traffic encapsulated in several commercial VPN tools, including PandaVPN, VPN-Super, X-VPN, and others. 

As shown in \cref{fig:acc-huawei}, NetTrans achieves the highest accuracy of 94.5\%, with NetMamba ranking second. Additionally, ET-BERT performs significantly worse than all other pre-trained models, underscoring the importance of header features for accurate VPN traffic classification.

\subsubsection{Comparison with Other NetX Variants on Public Datasets\label{sec:eval-netx-variants}}

\begin{table*}[htpb]
    \centering
    \caption{Comparison against Other Architectures in \name{} Backbone}
    \begin{tabular}{c|cc|cc|cc|cc|cc|cc}
    \toprule
    \multirow{2}{*}{Method} & \multicolumn{2}{c|}{Params(M)} & \multicolumn{2}{c|}{CipherSpectrum} & \multicolumn{2}{c|}{CSTNET-TLS1.3} & \multicolumn{2}{c|}{CICIoT2022} & \multicolumn{2}{c|}{ISCXVPN2016} & \multicolumn{2}{c}{USTC-TFC2016}\\
    \cmidrule(lr){2-13} \noalign{\vskip 0.5mm}
    & PT & FT & AC & F1 & AC & F1 & AC & F1 & AC & F1 & AC & F1 \\
    \midrule
    NetTransV & 2.3 & 1.9 &
    \underline{0.8504} & \underline{0.8535} & \cellcolor{lightgray}0.7848 & \cellcolor{lightgray}0.7820 & \underline{0.9731} & \underline{0.9731} & \underline{0.9283} & \underline{0.9284} & \cellcolor{lightgray}0.9825 & \cellcolor{lightgray}0.9825 \\
    NetTransL & 2.1 & 1.7 &
    0.3754 & 0.4567 & 0.2933 & 0.3725 & 0.7714 & 0.7839 & 0.7878 & 0.8142 & 0.7503 & 0.7918 \\
    \textbf{NetTrans} & 3.1 & 2.4 &
    \cellcolor{lightgray}0.8543 & \cellcolor{lightgray}0.8548 & \underline{0.7654} & \underline{0.7619} & \cellcolor{lightgray}0.9769 & \cellcolor{lightgray}0.9769 & \cellcolor{lightgray}0.9411 & \cellcolor{lightgray}0.9408 & \underline{0.9823} & \underline{0.9822} \\
    \midrule
    NetMambaB & 2.4 & 2.1 & \cellcolor{lightgray}0.9095 & \cellcolor{lightgray}0.9094 & \underline{0.7728} & \underline{0.7711} & \underline{0.9741} & \underline{0.9740} & \underline{0.9352} & \underline{0.9356} & \cellcolor{lightgray}0.9833 & \cellcolor{lightgray}0.9833 \\
    NetMambaC & 32.7 & 28.9 & 0.8310 & 0.8314 & 0.7327 & 0.7359 & 0.8838 & 0.8825 & 0.8890 & 0.8900 & 0.9370 & 0.9369 \\
    \textbf{NetMamba} & 2.2 & 1.9 &
    \underline{0.8779} & \underline{0.8783} & \cellcolor{lightgray}0.7755 & \cellcolor{lightgray}0.7728 & \cellcolor{lightgray}0.9779 & \cellcolor{lightgray}0.9779 & \cellcolor{lightgray}0.9401 & \cellcolor{lightgray}0.9401 & \underline{0.9743} & \underline{0.9740} \\
    \bottomrule
    \end{tabular}
    \label{tb:acc-variant}
\end{table*}

To assess the superiority of NetMamba and NetTrans blocks, we compare them to Mamba-based variants(i.e. NetMambaB and NetMambaC) and Transformer-based variants(i.e. NetTransV and NetTransL). 

As shown in \cref{tb:acc-variant}, among the Mamba-based variants, NetMamba achieves comparable accuracy to NetMambaB, while significantly outperforming NetMambaC. This indicates that a unidirectional scan is sufficient for aggregating sequential network traffic data, in contrast to the bidirectional scan of NetMambaB or the hierarchical scan of NetMambaC. Moreover, as shown in \Cref{sec:inference-efficiency}, incorporating these complex Mamba blocks introduces additional computational and memory overheads due to extra scan passes, ultimately reducing efficiency.

Regarding Transformer-based methods, NetTrans consistently outperforms both NetTransL and either surpasses or closely matches NetTransV in most cases. The performance gain over NetTransV is primarily attributed to the GeGLU-activated FFN, while the gain over NetTransL comes from the optimized quadratic attention mechanism.

\subsection{Inference Efficiency Evaluation\label{sec:inference-efficiency}}
\begin{figure*}[htpb]
    \centering
    \begin{minipage}{0.65\textwidth}
        \centering
        \includegraphics[scale=0.6]{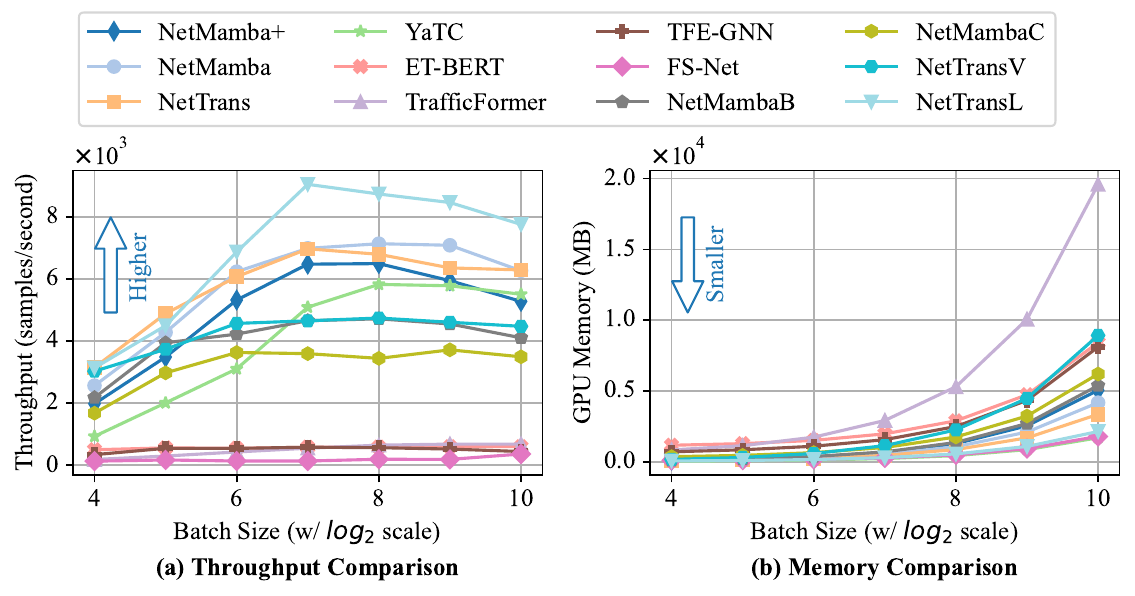}
        \caption{The Inference Speed and GPU Memory Comparison}
        \label{fig:efficiency}
    \end{minipage}
    \begin{minipage}{0.3\textwidth}
        \centering
        \includegraphics[scale=0.6]{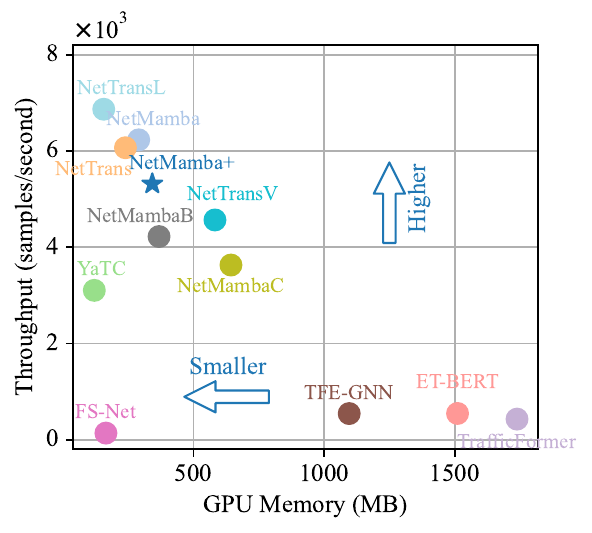}
        \caption{The Inference Efficiency Comparison on Fine-tuning Batch Size}
        \label{fig:efficiency-bar}
    \end{minipage}
\end{figure*}

To evaluate the inference efficiency of NetMamba and NetTrans, we conducted experiments comparing their speed and GPU memory consumption with existing deep learning methods and other \name{} variants. Speed is measured as the number of traffic data samples processed by the model per second: packets for ET-BERT and flows for the others.

\subsubsection{Throughput Comparison}
As shown in \cref{fig:efficiency}(a), both NetMamba and NetTrans exhibit similar inference throughputs across various input batch sizes, while NetMamba+ obtains lower throughput and consumes a little more memory due to processing extra size and interval tokens. Compared to existing SOTA methods, NetMamba+ achieves throughput improvements ranging from 0.96 to 47.4 times. 
This advantage is particularly notable due to the substantial model parameters and inefficient model architecture design present in models such as ET-BERT, TrafficFormer, TFE-GNN, and FS-Net.

Even when compared with models possessing similar parameter counts, NetMamba and NetTrans continue to outperform YaTC and NetTransV, both of which employ the vanilla Transformer with quadratic complexity. This superior performance of NetMamba can be attributed to its linear computational complexity, hardware-based parallel scan, and kernel fusion, while NetTrans benefits from its IO-aware tiling and recomputation techniques. 
Additionally, the bidirectional or hierarchical scan significantly reduces the inference speed of NetMambaB and NetMambaC.

Finally, NetTransL achieves the highest inference throughput among all models due to the inherently linear time and memory complexity of its attention mechanism, independent of processor-level parallelization or hardware-specific optimization. However, this efficiency comes at the expense of unstable classification performance, caused by the over-compression of attention scores.

\subsubsection{Memory Comparison}
In \cref{fig:efficiency}(b), NetTrans, NetMamba and NetMamba+ demonstrate lower GPU memory consumption than most models, except for FS-Net, YaTC and NetTransL, when using large batch sizes. 
FS-Net's reliance on RNNs, which require linear memory relative to sequence length, reduces memory costs but results in slower inference and poorer classification performance.
YaTC reduces memory usage by shortening input sequence length through a model forward trick, while NetTransL benefits from its linear complexity.

Compared to other baselines, our models achieve improved memory efficiency primarily by customizing GPU operators that minimize the storage of extensive intermediate states and perform recomputation during the backward pass.

When the input batch size is set to 64 (the value used in fine-tuning), as depicted in Figure \ref{fig:efficiency-bar}, NetMamba+ exhibits a 1.7x speed improvement over the best existing SOTA method, YaTC. Apart from FS-Net, YaTC, and NetTransL, our models surpass other methods in terms of GPU memory utilization. In summary, both our models achieve high inference speeds while maintaining comparably low memory usage compared to all deep learning methods.

\begin{table*}[htpb]
    \caption{Ablation Results on Raw Bytes}
    \centering
    \begin{threeparttable}
    \begin{tabular}{c|cc|cc|cc|cc|cc}
    \toprule
    \multirow{2}{*}{Method} & \multicolumn{2}{c|}{CipherSpectrum} & \multicolumn{2}{c|}{CSTNET-TLS1.3} & \multicolumn{2}{c|}{CICIoT2022} & \multicolumn{2}{c|}{ISCXVPN2016} & \multicolumn{2}{c}{USTC-TFC2016}\\
    \cmidrule(lr){2-11} \noalign{\vskip 0.5mm}
    & AC & F1 & AC & F1 & AC & F1 & AC & F1  & AC & F1 \\
    \midrule
    \textbf{NetMamba (default)} &
    \cellcolor{lightgray}0.8779 & \cellcolor{lightgray}0.8783 & \underline{0.7755} & \underline{0.7728} & \cellcolor{lightgray}0.9779 & \cellcolor{lightgray}0.9779 & 0.9401 & 0.9401 & 0.9743 & 0.9740 \\
    \midrule
    w/o Header & 0.7293 & 0.7381 & 0.6219 & 0.6279 & 0.5351 & 0.5591 & 0.4921 & 0.4524 & 0.4993 & 0.5581 \\
    w/o Payload & \underline{0.8554} & 0.8550 & \cellcolor{lightgray}0.7774 & \cellcolor{lightgray}0.7740 & 0.9635 & 0.9634 & \cellcolor{lightgray}0.9440 & \cellcolor{lightgray}0.9443 & \underline{0.9783} & \underline{0.9782} \\
    w/o Stride Cutting\tnote{1} & 0.8552 & \underline{0.8560} & 0.7609 & 0.7579 & \underline{0.9712} & \underline{0.9712} & \underline{0.9430} & \underline{0.9433} & \cellcolor{lightgray}0.9808 & \cellcolor{lightgray}0.9808 \\
    w/o Pre-training & 0.8529 & 0.8520 & 0.7108 & 0.7051 & 0.9549 & 0.9548 & 0.8861 & 0.8868 & 0.9770 & 0.9771 \\
    \bottomrule
    \end{tabular}
    \begin{tablenotes}
        \item[1] Changed the 1-dimensional stride cutting to 2-dimensional patch splitting.
    \end{tablenotes}
    \end{threeparttable}
    \label{tb:ablation}
\end{table*}

\subsection{Ablation on Raw Bytes \label{sec:ab-study}}
To further validate both our pre-training design and the traffic representation scheme on raw byte features, we conducted ablation studies of NetMamba to assess the contribution of each component across five public datasets. The results are presented in \cref{tb:ablation}.

\begin{figure}[htpb]
    \centering
    \begin{minipage}{0.2\textwidth}
        \centering
        \includegraphics[scale=0.16]{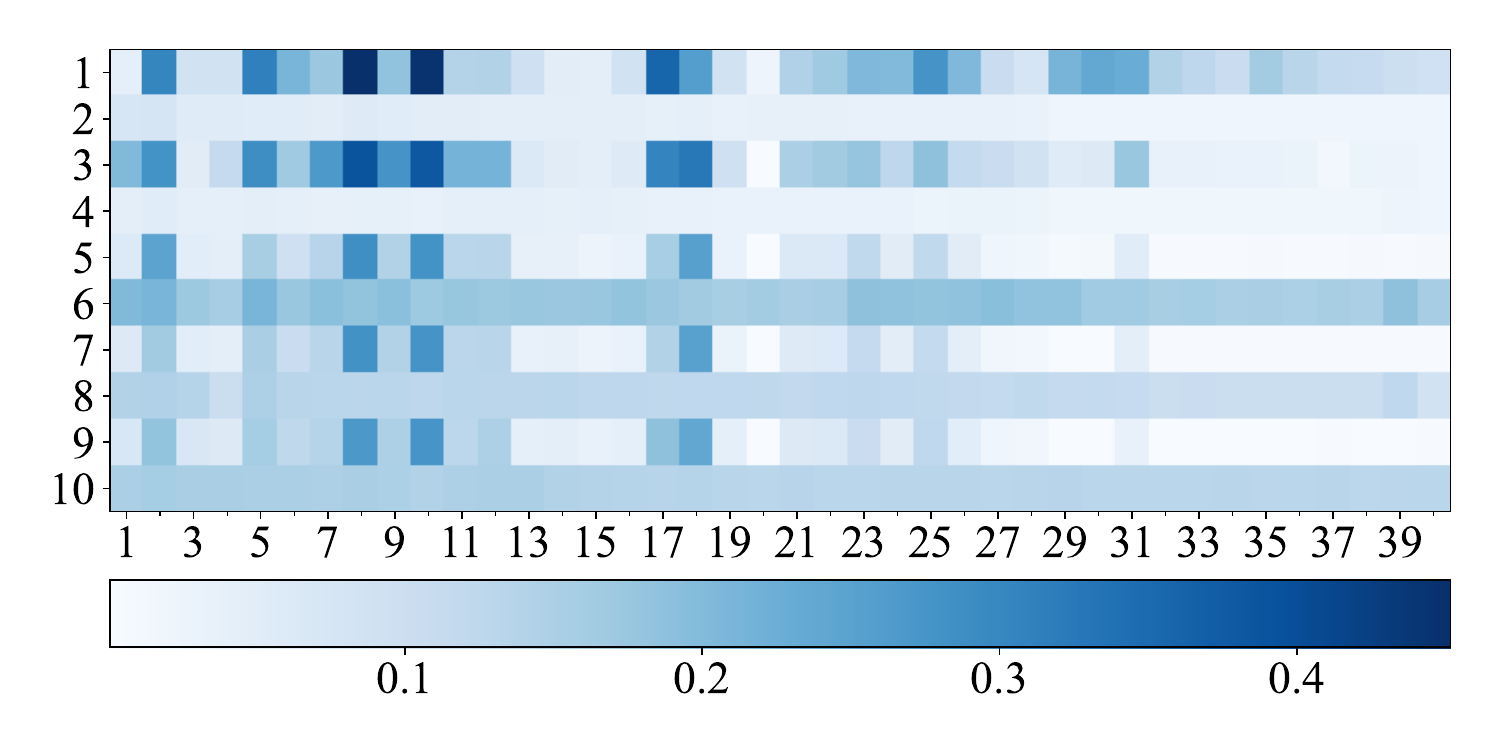}
        \captionsetup{font=footnotesize}
        \caption*{(a) CICIoT2022}
        \label{fig:hm-ciciot}
    \end{minipage}
    \hspace{0.2cm} 
    \begin{minipage}{0.2\textwidth}
        \centering
        \includegraphics[scale=0.16]{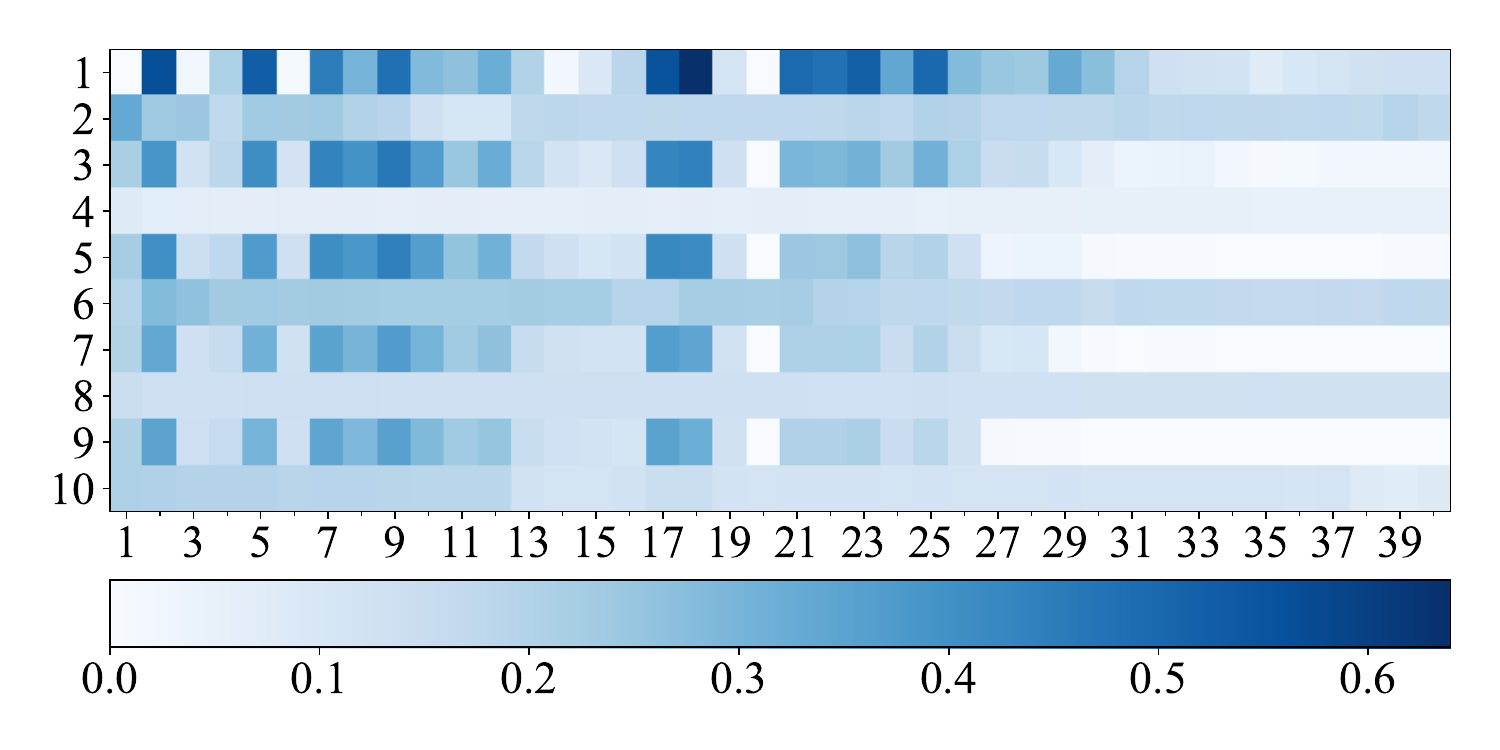}
        \captionsetup{font=footnotesize}
        \caption*{(b) USTC-TFC2016}
        \label{fig:hm-ustc-tfc}
    \end{minipage}
    \captionsetup{font=normal}
    \caption{Mean AMI Scores for 2-Byte Strides on the CICIoT2022 and USTC-TFC2016 Datasets. The strides are arranged from left to right and from top to bottom. The strides of the $i$-th packet are placed in the $(2i-1)$-th and $(2i)$-th rows, with the former representing header strides and the latter representing payload bytes, for $i \in \{1, \dots, 5\}$.}
    \label{fig:heatmap}
\end{figure}

We first remove all header bytes from each packet and observe substantial performance degradation, with accuracy decreasing by 14.86–47.50\%. To further analyze feature importance, we compute the adjusted mutual information (AMI)~\cite{van2020flowprint} in a model-agnostic manner to evaluate the contribution of strides at different positions to traffic classification.

As shown in \Cref{fig:heatmap}, the AMI scores of certain packet strides are notably higher than those of payload bytes, highlighting their critical role in traffic classification. Apart from IP addresses (excluded prior to model input), header fields such as \emph{total length}(indexed as 2 on the x-axis), protocol(indexed as 5 on the x-axis), \emph{TCP flags}(indexed as 17 on the x-axis), and \emph{TCP window size}(indexed as 18 on the x-axis) are of significant importance. The contribution of these header features aligns with findings from previous studies\cite{liu2019fs,madhukar2006longitudinal,fu2021realtime}. 

When packet payloads are removed, we observe unstable ablation results, with performance drops in two cases but improvements in three. This indicates the inconsistent contribution of payload features and suggests that excluding them may be preferable when efficiency is prioritized.

Replacing coarse-grained two-dimensional patch splitting with fine-grained one-dimensional stride cutting leads to a maximum accuracy reduction of only 2.27\% on the CipherSpectrum dataset. Compared to patch splitting, stride cutting offers two main advantages. First, a square patch groups vertically adjacent but semantically unrelated bytes into a single token, introducing biased correlations based on the structure of the image-like byte matrix. Second, stride cutting with a 2-byte granularity aligns better with the length of key packet fields, such as total length, port number, and TCP window size, thereby preserving more fine-grained protocol information.

Finally, in the absence of pre-training, performance declines are observed in four settings, underscoring the importance of generic domain knowledge acquired during self-supervised pre-training.

\subsection{Ablation on Multimodal Features \label{sec:ab-mmf}}

\begin{table*}[htpb]
    \caption{Ablation Results on Multimodal Features}
    \centering
    \begin{threeparttable}
    \begin{tabular}{c|cc|cc|cc|cc|cc}
    \toprule
    \multirow{2}{*}{Input feature} & \multicolumn{2}{c|}{CipherSpectrum} & \multicolumn{2}{c|}{CSTNET-TLS1.3} & \multicolumn{2}{c|}{CICIoT2022} & \multicolumn{2}{c|}{ISCXVPN2016} & \multicolumn{2}{c}{USTC-TFC2016}\\
    \cmidrule(lr){2-11} \noalign{\vskip 0.5mm}
    & AC & F1 & AC & F1 & AC & F1 & AC & F1  & AC & F1 \\
    \midrule
    Byte Only &
    0.8779 & 0.8783 & 0.7755 & 0.7728 & \cellcolor{lightgray}0.9769 & \cellcolor{lightgray}0.9779 & \underline{0.9401} & \underline{0.9401} & \underline{0.9743} & \underline{0.9740} \\
    Size Only & \underline{0.8906} & \underline{0.8916} & \underline{0.8112} & \underline{0.8094} & 0.8377 & 0.8381 & 0.7554 & 0.7479 & 0.7460 & 0.7513 \\
    Interval Only & 0.7556 & 0.7553 & 0.5076 & 0.5006 & 0.5908 & 0.5890 & 0.5029 & 0.4825 & 0.4243 & 0.4129 \\ \midrule
    \textbf{All (NetMamba+)} & \cellcolor{lightgray}0.9652 & \cellcolor{lightgray}0.9652 & \cellcolor{lightgray}0.8498 & \cellcolor{lightgray}0.8498 & \underline{0.9750} & \underline{0.9750} & \cellcolor{lightgray}0.9460 & \cellcolor{lightgray}0.9460 & \cellcolor{lightgray}0.9765 & \cellcolor{lightgray}0.9765 \\
    \bottomrule
    \end{tabular}
    \end{threeparttable}
    \label{tb:mmf}
\end{table*}
Results in \Cref{tb:acc-sota} highlight the importance of sequence features such as packet sizes and time intervals. To examine their joint effect with raw bytes, we integrate all three modalities through an early-fusion approach. 
As shown in \Cref{tb:mmf}, raw bytes and packet sizes individually dominate performance on different datasets. By combining detailed packet content from raw bytes, spatial patterns from size sequences, and temporal dynamics from interval sequences, NetTrans with multimodal features achieves the best performance on four datasets. 

This multimodal fusion not only enriches the input representation and enhances overall performance but also helps mitigate the risk of overfitting inherent to raw-byte features alone.

\subsection{Ablation on LDA Fine-tuning\label{sec:lda-eval}}

\begin{table}[htpb]
    \caption{Ablation Results on LDA Fine-tuning}
    \centering
    \begin{tabular}{c|c|cc|cc}
        \toprule
        \multirow{2}{*}{Dataset} &
        \multirow{2}{*}{Fine-tuning} & \multicolumn{2}{c|}{NetTrans} & \multicolumn{2}{c}{NetMamba} \\
        \cmidrule(lr){3-6} \noalign{\vskip 0.5mm}
        & & AC & F1 & AC & F1 \\
        \midrule
        \multirow{2}{*}{Huawei-VPN} & w/o LDA & 0.9366 & 0.9367 & 0.9311 & 0.9315 \\
        & w/ LDA & \cellcolor{lightgray}0.9445 & \cellcolor{lightgray}0.9450 & \cellcolor{lightgray}0.9361 & \cellcolor{lightgray}0.9361 \\
        \midrule
        \multirow{2}{*}{CrossNet2021-A} & w/o LDA & 0.9029 & 0.9028 & 0.9062 & 0.9064 \\
        & w/ LDA & \cellcolor{lightgray}0.9040 & \cellcolor{lightgray}0.9041 & \cellcolor{lightgray}0.9161 & \cellcolor{lightgray}0.9162 \\
        \midrule
        \multirow{2}{*}{CP-Android} & w/o LDA & \cellcolor{lightgray}0.7456 & \cellcolor{lightgray}0.7445 & 0.7648 & 0.7614 \\
        & w/ LDA & 0.7392 & 0.7373 & \cellcolor{lightgray}0.7823 & \cellcolor{lightgray}0.7803 \\
        \midrule
        \multirow{2}{*}{CP-iOS} & w/o LDA & 0.6382 & 0.6340 & 0.6511 & 0.6462 \\
        & w/ LDA & \cellcolor{lightgray}0.6670 & \cellcolor{lightgray}0.6637 & \cellcolor{lightgray}0.6700 & \cellcolor{lightgray}0.6702 \\
        \midrule
        \multirow{2}{*}{DataCon2021-p1} & w/o LDA & 0.8682 & 0.8672 & 0.8811 & 0.8821 \\
        & w/ LDA & \cellcolor{lightgray}0.8786 & \cellcolor{lightgray}0.8784 & \cellcolor{lightgray}0.8992 & \cellcolor{lightgray}0.8996 \\
        \bottomrule
    \end{tabular}
    \label{tb:lda-huawei}
\end{table}

\begin{figure}[htpb]
    \centering
    \begin{minipage}{0.2\textwidth}
        \centering
        \includegraphics[scale=0.27]{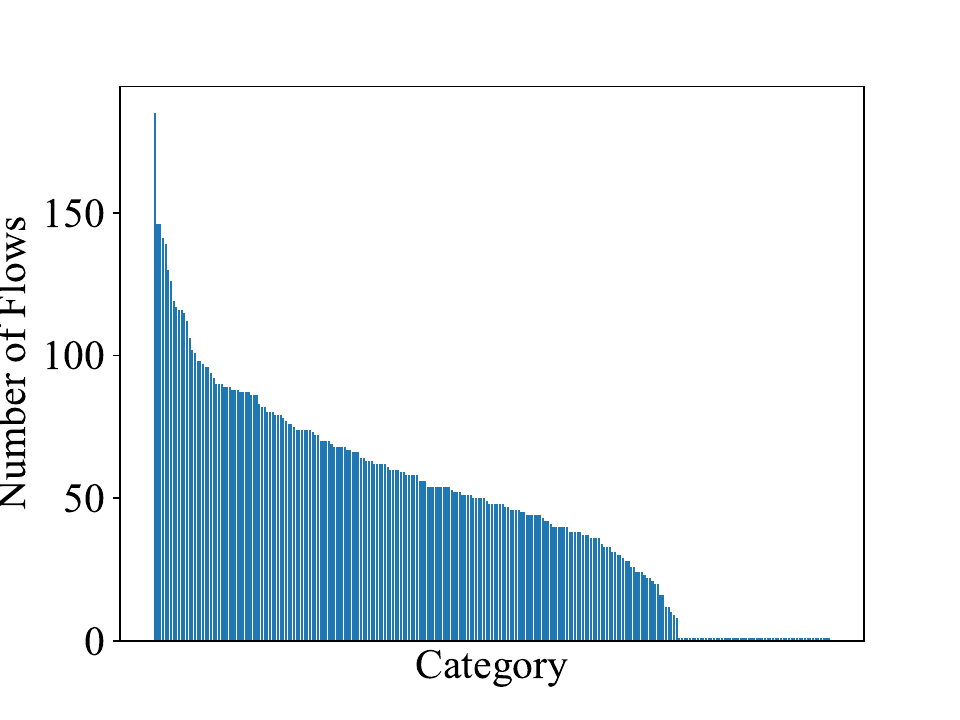}
        \captionsetup{font=footnotesize}
        \caption*{(a) Before Filtering}
        \label{fig:flow-distri-raw}
    \end{minipage}
    \hspace{0.2cm} 
    \begin{minipage}{0.2\textwidth}
        \centering
        \includegraphics[scale=0.27]{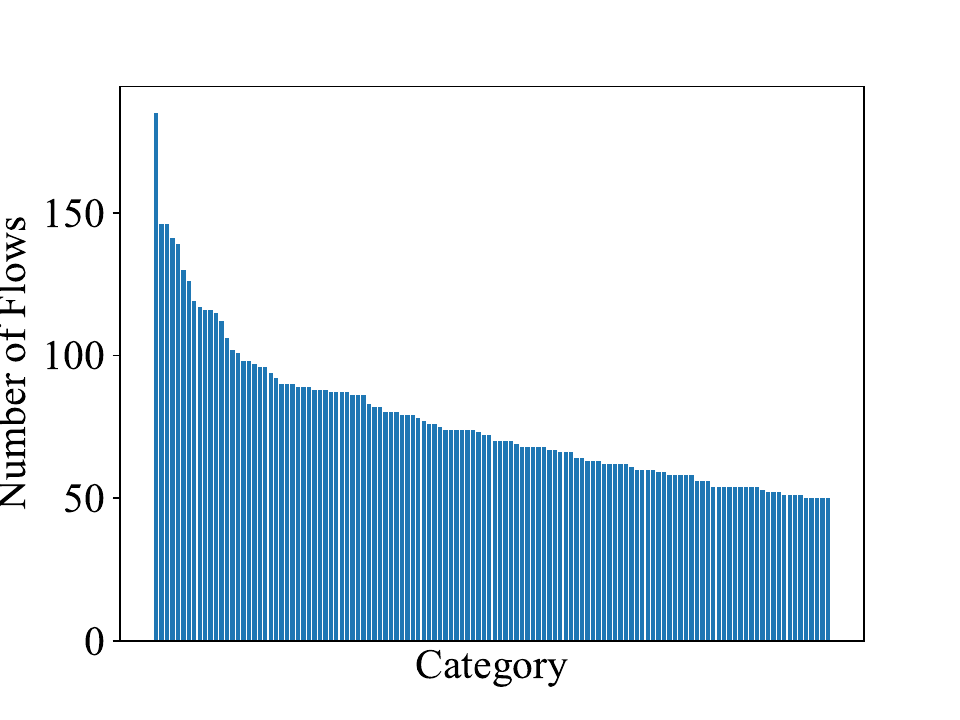}
        \captionsetup{font=footnotesize}
        \caption*{(b) After Filtering}
        \label{fig:flow-distri-sampled}
    \end{minipage}
    \captionsetup{font=normal}
    \caption{Comparison of CP-iOS Flow Distribution 
    }
    \label{fig:flow-distri}
\end{figure}

The label distribution-aware~(LDA) fine-tuning approach introduces a novel cross-entropy loss that assigns higher weights and enforces larger margins for minority classes in training datasets, aiming to improve performance on long-tailed datasets. Data imbalance is prevalent in both public and real-world datasets. As shown in \Cref{fig:flow-distri}(a), a small portion of categories in the original CP-iOS dataset have flow numbers exceeding 100, while nearly half of the categories have fewer than 50, and almost a quarter have close to zero. After filtering out categories with few samples, the imbalance is somewhat mitigated, as shown in \Cref{fig:flow-distri}(b), but the disparity remains.

This motivates the application of LDA fine-tuning on both filtered public datasets and the unfiltered real-world dataset. When replacing our proposed LDA cross-entropy loss with a standard one, we observed accuracy drops of up to 2.88\% on public datasets and 0.79\% on Huawei-VPN dataset, as shown in \Cref{tb:lda-huawei}. These results confirm that LDA fine-tuning enhances the ability of both NetMamba and NetTrans to adapt to imbalanced datasets.

\subsection{Few-Shot Evaluation}
\begin{figure*}[h]
\centering
\includegraphics[scale=0.6]{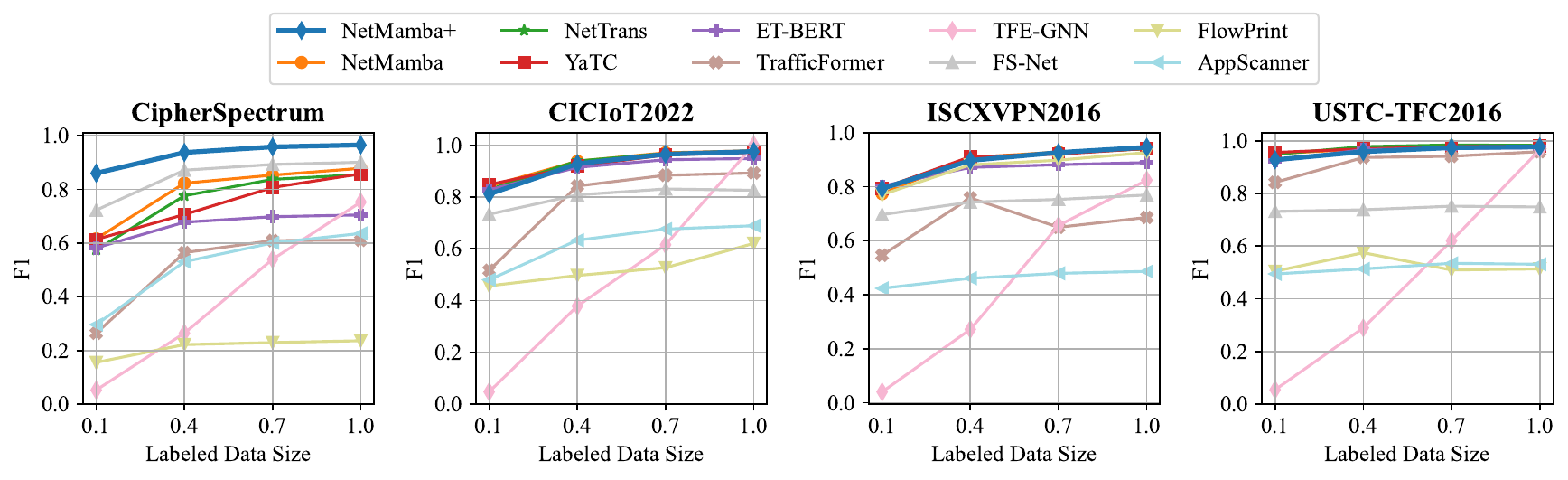}
\caption{The Performance Comparison on Few-Shot Settings}
\label{fig:few-shot}
\end{figure*}

To validate the robustness of our models, we conduct few-shot evaluations on four datasets, with labeled data size set to 10\%, 40\%, 70\%, and 100\% of the full training set (comprising 80\% of the total data). 
As shown in \cref{fig:few-shot}, the four pre-trained models—NetMamba+, NetMamba, NetTrans, YaTC, and ET-BERT—generally outperform other supervised methods under few-shot settings, whereas TrafficFormer exhibits relatively weak performance among pre-trained approaches.
Conventional machine learning methods such as FlowPrint and AppScanner demonstrate some robustness to limited labeled data, but their performance varies substantially across datasets. Among supervised deep learning models, FS-Net is less sensitive to training data size but still underperforms compared with pre-trained methods on most datasets. Although TFE-GNN performs competitively with pre-trained models when trained on the full dataset, its accuracy declines sharply under few-shot scenarios.

These results confirm that pre-trained models possess superior robustness and generalization, owing to their ability to learn high-quality traffic representations from large amounts of unlabeled data and thereby reduce reliance on labeled samples. In particular, NetMamba+ demonstrates best robustness—superior to vanilla Transformer-based models—and proves highly effective in handling classification tasks with limited encrypted traffic data.

\subsection{Out-of-Distribution Evaluation}
\begin{table*}[htpb]
    \caption{Out-of-Distribution Evaluation Results}
    \centering
    \begin{threeparttable}
    \begin{tabular}{c|cc|cc|cc|cc}
    \toprule
    \multirow{2}{*}{Model} & \multicolumn{2}{c|}{Unknown Application$^{\P}$} & \multicolumn{2}{c|}{Unknown Attack$^{*}$} & \multicolumn{2}{c|}{Unknown VPN$^{\dagger}$} & \multicolumn{2}{c}{Unknown Malware$^{\ddagger}$}\\
    \cmidrule(lr){2-9} \noalign{\vskip 0.5mm}
    & AUROC($\uparrow$) & FPR95($\downarrow$) & AUROC($\uparrow$)  & FPR95($\downarrow$) & AUROC($\uparrow$) & FPR95($\downarrow$) & AUROC($\uparrow$)  & FPR95($\downarrow$) \\
    \midrule
    NetMamba & 0.8817 & 0.2744 & 0.9061 & 0.2205 & 0.9242 & 0.1946 & 0.9245 & 0.1765 \\
    NetMamba+ & 0.9455 & 0.3670 & 0.9825 & 0.0463 & 0.9720 & 0.0715 & 0.9668 & 0.1072 \\
    \bottomrule
    \end{tabular}
    \begin{tablenotes}
        \item $^{\P}$OOD dataset: CSTNET-TLS1.3; $^{*}$ OOD dataset: CICIoT2022; $^{\dagger}$OOD dataset: ISCXVPN2016; $^{\ddagger}$OOD dataset: USTC-TFC2016
    \end{tablenotes}
    \end{threeparttable}
    \label{tb:ood}
\end{table*}
To further evaluate NetMamba+’s capability in identifying previously unseen traffic categories, we introduce four out-of-distribution (OOD) detection tasks: unknown application, unknown attack, unknown VPN, and unknown malware. For each task, CipherSpectrum is used as the sole in-distribution (ID) dataset. 

For simplicity, OOD samples are detected by computing the temperature-scaled entropy of the predicted probability vectors. Formally, given prediction logits $\mathbf{z} \in \mathbb{R}^{\mathtt{C}}$ and temperature $\tau$, OOD detection is performed as follows:
$$
\hat{y} = 
\begin{cases}
0, & \text{if } \sum_{i=1}^{\mathtt{C}}p_i \log p_i \ge s \\
1, & \text{otherwise}
\end{cases}
$$ where $p_i = \frac{\text{exp}(z_i/\tau)}{\sum_{j}\text{exp}(z_j/\tau)}$ denotes the predicted probability for the $i$-th category, and $s$ is a pre-defined threshold.

As reported in \Cref{tb:ood}, by selecting an appropriate temperature, NetMamba+ achieves excellent OOD detection performance across all four tasks, reaching a maximum AUROC of 0.9825 and a minimum FPR95 of 0.0463. Although inferior to NetMamba+, the raw-byte-based NetMamba still exhibits reasonably strong OOD detection, with AUROCs exceeding 0.9 in three tasks. These results indicate that our models are capable of effectively identifying novel traffic categories in real-world deployment scenarios.

\subsection{Real-World Deployment}
\begin{figure}[htpb]
    \centering
    \begin{minipage}{0.2\textwidth}
        \centering
        \includegraphics[scale=0.25]{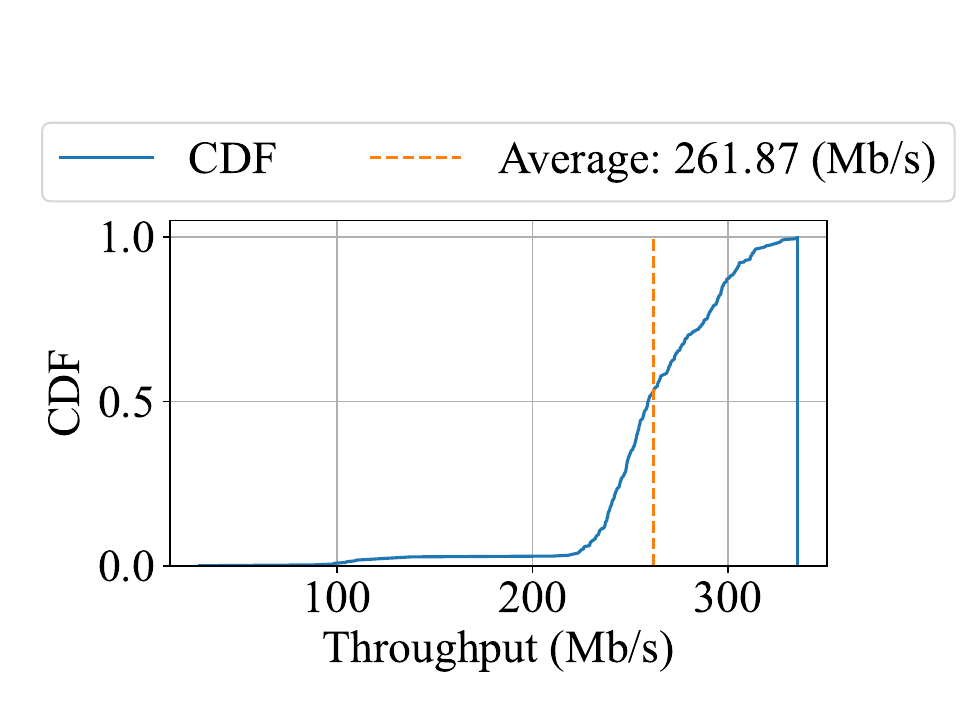}
        \captionsetup{font=footnotesize}
        \caption*{(a) Batch Throughput}
        \label{fig:batch-throughput}
    \end{minipage}
    \hspace{0.2cm} 
    \begin{minipage}{0.2\textwidth}
        \centering
        \includegraphics[scale=0.25]{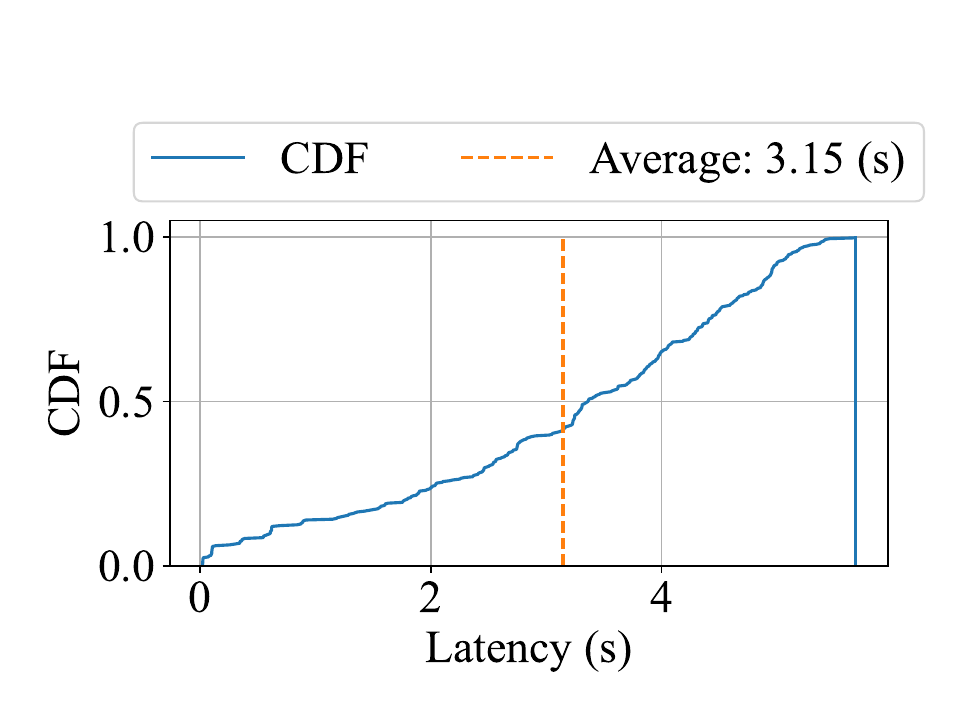}
        \captionsetup{font=footnotesize}
        \caption*{(b) Batch Latency}
        \label{fig:batch-latency}
    \end{minipage}
    \captionsetup{font=normal}
    \caption{Batch Inference Throughput and Latency of NetMamba}
    \label{fig:throughput}
\end{figure}

We deploy the NetMamba-based online system, as depicted in \Cref{sec:online-sys}, in real-world environments. Network traffic is generated by a commercial network tester, then captured and processed by our online system. This system runs on a Ubuntu 20.04 server with an Intel(R) Xeon(R) Platinum 8375C CPU (2.90 GHz, 32 cores), 512 GB memory, an NVIDIA Mellanox ConnectX-5 NIC that supports DPDK, and an A30 Tensor Core GPU (24 GB). We allocate one physical core for traffic capturing using DPDK 24.03, one core for feature processing by a C++ process, and one core for NetMamba inference by a Python process. 

We generate a new batch of traffic flows every $W_g=3$ seconds and remove outdated flow entries every $W_r=10$ seconds. After reading data from shared memory, NetMamba performs classification on dynamic batches, and the corresponding cumulative distribution function~(CDF) of inference throughput and latency is shown in \cref{fig:throughput}. The batch throughput for online inference varies between 29.85 Mb/s and 335.21 Mb/s, with an average of 261.87 Mb/s. The batch latency ranges from 0.02 seconds to 5.68 seconds, with an average of 3.15 seconds. 

\subsection{Discussion}
\subsubsection{Model Generalization Ability}
A model demonstrates strong generalization ability if it achieves competitive performance on test data whose distribution differs from that of the training set. To assess the generalization of \name{}, we sort all flows by the timestamp of their first packet in ascending order. The earlier flows are used for training and validation, while the later flows form the test set. After evaluation, the generalization performance of \name{} varies across datasets. On the CipherSpectrum dataset, we observe an accuracy of 0.9610, with an drop of 0.42\%. For CSTNET-TLS1.3, the accuracy drop is 8.47\%. In summary, \name{} exhibit performance degradation to some extent under distribution shift settings. Enhancing the model’s robustness under such distribution shifts remains an important direction for future work.

\subsubsection{Training Efficiency}
Although our experiments mainly compare inference efficiency, \name{} also demonstrates superior training efficiency over existing models. For Transformer-based methods, the advantage comes from reduced model complexity and fewer parameters. For RNN- and GNN-based methods, the gains are due to parallelism and smaller parameter counts. Since inference efficiency has a more direct impact on deployment costs, we primarily report inference results in this work.
\section{Conclusion}
In this paper, we introduce \name{}, a novel framework of pre-trained models featuring efficient underlying architectures, a comprehensive traffic representation scheme, and a label distribution-aware fine-tuning strategy.
\name{} enhances model efficiency by optimizing the unidirectional Mamba and Flash Attention-based Transformer architectures. To improve performance, we develop a multimodal traffic representation scheme and introduce label distribution-aware fine-tuning.
Furthermore, we implement an online traffic classification system for \name{}, demonstrating its practical applicability for real-world deployment.
Evaluation experiments on massive datasets demonstrate the superior effectiveness, efficiency, and robustness of \name{}.
Beyond classical traffic classification tasks, the comprehensive representation scheme and refined model design enable \name{} to address broader tasks within the network domain, such as quality of service prediction and network performance prediction. 

a
\bibliographystyle{IEEEtran}
\bibliography{reference}

\vspace{-32pt}
\begin{IEEEbiographynophoto}
{Tongze Wang}
received his B.S. degree in Computer Science and Technology from Tsinghua University, Beijing, China, in 2023. He is currently pursuing the M.S. degree at the Institute for Network Sciences and Cyberspace, Tsinghua University. His research interests include machine learning, traffic analysis, and LLM for networking.
\end{IEEEbiographynophoto}

\vspace{-32pt}
\begin{IEEEbiographynophoto}
{Xiaohui Xie}(Member, IEEE)
received the B.E. and Ph.D. degrees in computer science and engineering from Tsinghua University, China, in 2016 and 2021, respectively. He is currently an Assistant Professor at the Computer Science Department, Tsinghua University.
His research interests include Network Anomaly Detection, Information Retrieval, and Artificial intelligence.
\end{IEEEbiographynophoto}

\vspace{-32pt}
\begin{IEEEbiographynophoto}
{Wenduo Wang}
received the B.S. degree from Tsinghua University, Beijing, China, in 2024, where he is currently pursuing the Ph.D. degree in the Department of Computer Science and Technology. His current research interests include network traffic analysis, network security, and deep learning.
\end{IEEEbiographynophoto}

\vspace{-32pt}
\begin{IEEEbiographynophoto}
{Chuyi Wang}
(Graduate Student Member, IEEE) received her B.E. degree in computer science and technology from Tsinghua University, Beijing, China, in 2025. She is currently pursuing the Ph.D degree at the Department of Computer Science and Technology, Tsinghua University. Her research interests include encrypted network traffic classification and AI agent communication.
\end{IEEEbiographynophoto}

\vspace{-32pt}
\begin{IEEEbiographynophoto}
{Jinzhou Liu}
received his B.S. degree in Software Engineering from Hebei Normal University, China, in 2020, and his M.S. degree from the Institute of Information Engineering, Chinese Academy of Sciences, Beijing, China, in 2023. He is currently an Assistant Engineer at Beijing Zhongguancun Laboratory. His research interests include network security and computer architecture.
\end{IEEEbiographynophoto}

\vspace{-32pt}
\begin{IEEEbiographynophoto}
{Boyan Huang}
is currently pursuing the bachelor’s degree with the School of Computer Science and Engineering, Central South University, China. His research interests include AI for networking and networking for AI. 
\end{IEEEbiographynophoto}

\vspace{-32pt}
\begin{IEEEbiographynophoto}
{Yannan Hu}
received the B.S. degree from Harbin Institute of Technology, Harbin, China in 2008 and Ph.D. degree from Beijing University of Posts and Telecommunications, Beijing, China in 2015. He is currently an Associate Researcher at Beijing Zhongguancun Laboratory. His research interests include network security, network architecture and machine learning.
\end{IEEEbiographynophoto}

\vspace{-32pt}
\begin{IEEEbiographynophoto}
{Youjian Zhao}
received the B.S. degree from Tsinghua University, Beijing, China, in 1991, the M.S. degree from the Shenyang Institute of Computing Technology, Chinese Academy of Sciences, in 1995, and the Ph.D. degree in compute science from Northeastern University, China, in 1999. He is currently a Professor with the Computer Science and Technology Department, Tsinghua University. His research interests include high-speed Internet architecture, switching and routing, and anomaly detection for network data.
\end{IEEEbiographynophoto}

\vspace{-32pt}
\begin{IEEEbiographynophoto}
{Yong Cui}(Member, IEEE)
received the B.E. and Ph.D. degrees both on Computer Science and Engineering from Tsinghua University. He is currently a full professor at the Computer Science Department in Tsinghua University. He served or serves at the editorial boards on IEEE TPDS, IEEE TCC, IEEE Network and IEEE Internet Computing. He published over 100 papers with several Best Paper Awards and 10 Internet standard documents (RFC). His research interests include Internet architecture and data-driven network.
\end{IEEEbiographynophoto}


\end{document}